\documentclass[pdflatex,sn-mathphys]{sn-jnl}
\jyear{2023}%
\theoremstyle{thmstyleone}

\theoremstyle{thmstyletwo}

\theoremstyle{thmstylethree}

\usepackage{graphicx}
\usepackage{textcomp}
\usepackage{xcolor}
\usepackage{url}
\usepackage{multirow} 
\usepackage{tabularx}
\usepackage{lipsum}
\usepackage[caption=false]{subfig}

\begin{document}
\title[Improving Position Encoding of Transformers for MTSC]{Improving Position Encoding of Transformers for Multivariate Time Series Classification}


\author*[1]{\fnm{Navid} \sur{Mohammadi Foumani}}\email{navid.foumani@monash.edu.com}

\author[1]{\fnm{Chang Wei} \sur{Tan}}\email{chang.tan@monash.edu}
\author[1]{\fnm{Geoffrey I.} \sur{Webb}}\email{geoff.webb@monash.edu}
\author[1]{\fnm{Mahsa} \sur{Salehi}}\email{mahsa.salehi@monash.edu}

\affil*[1]{\orgdiv{Department of Data Science and Artificial Intelligence}, \orgname{Monash~University}, \orgaddress{ \city{Melbourne}, \state{VIC}, \country{Australia}}}

\abstract{
Transformers have demonstrated outstanding performance in many applications of deep learning. When applied to time series data, transformers require effective position encoding to capture the ordering of the time series data. The efficacy of position encoding  in time series analysis is not well-studied and remains controversial, e.g., whether it is better to inject absolute position encoding or relative position encoding, or a combination of them. In order to clarify this, we first review  existing absolute and relative position encoding methods when applied in time series classification. We then proposed a new absolute position encoding method dedicated to time series data called time Absolute Position Encoding (tAPE). Our new method incorporates the series length and input embedding dimension in absolute position encoding.  Additionally, we propose computationally Efficient implementation of Relative Position Encoding (eRPE) to improve generalisability for time series. We then propose a novel multivariate time series classification (MTSC) model combining tAPE/eRPE and convolution-based input encoding named ConvTran to improve the position and data embedding of time series data. The proposed absolute and relative position encoding methods are simple and efficient. They can be easily integrated into transformer blocks and used for downstream tasks such as forecasting, extrinsic regression, and anomaly detection. Extensive experiments on 32 multivariate time-series datasets show that our model is significantly more accurate than state-of-the-art convolution and transformer-based models.
Code and models are open-sourced at \url{https://github.com/Navidfoumani/ConvTran}.

}

\maketitle

\section{Introduction}\label{sec1}

A time series is a time-dependent quantity recorded over time. Time series data can be univariate, where only a sequence of values for one variable is collected; or multivariate, where data are collected on multiple variables. There are many applications that require time series analysis, such as human activity recognition \cite{lockhart2011design}, diagnosis based on electrocardiogram (ECG), electroencephalogram (EEG), and systems monitoring problems \cite{bagnall2018uea}. Many of these applications are inherently multivariate in nature -- various sensors are used to measure human's activities; EEGs use a set of electrodes (channels) to measure brain signals at different locations of the brain. Hence, multivariate time-series analysis methods such as classification and segmentation are of great current interest \cite{bagnall2017great,fawaz2019deep,ruiz2020great}.

Convolutional neural networks (CNNs) have been widely employed in time series classification \cite{fawaz2019deep,ruiz2020great}. Many studies have shown that convolution layers tend to have strong generalization with fast convergence due to their strong inductive bias \cite{dai2021coatnet}. While CNN-based models are excellent for capturing local temporal/spatial correlations, these models cannot effectively capture and utilize long-range dependencies. Also, they only consider the local order of data points in a time series rather than the order of all data points globally. Due to this, many recent studies have used recurrent neural networks (RNN) such as LSTMs to capture this information \cite{karim2019multivariate}. However, RNN-based models are computationally expensive, and their capability in capturing long-range dependencies are limited \cite{vaswani2017attention, hao2020new}.

On the other hand, \textit{attention models} can capture long-range dependencies, and their broader receptive fields provide more contextual information, which can improve the models' learning capacity. Not surprisingly, with the success of attention models in natural language processing \cite{vaswani2017attention,devlin2018bert}, many previous studies have attempted to bring the power of attention models into other domains such as computer vision \cite{dosovitskiy2020image} and time series analysis \cite{hao2020new,zerveas2021transformer,kostas2021bendr}.

The transformer's core is self-attention \cite{vaswani2017attention}, which is capable of modeling the relationship of input time series. Self-attention, however, has a limitation — it cannot capture the ordering of input series.
Hence, adding explicit representations of position information is especially important for the attention since the model is otherwise entirely invariant to input order, which is undesirable for modeling sequential data. This limitation is even worse in time series data since, unlike image and text, which use Word2Vec-like embedding, time series data has less informative data context. 

There are two main methods for encoding positional information in transformers: absolute and relative. Absolute methods, such as those used in~\cite{vaswani2017attention,devlin2018bert}, assign a unique encoding vector to each position in the input sequence based on its absolute position in the sequence. These encoding vectors are combined with the input encoding to provide positional information to the model. On the other hand, relative methods~\cite{shaw2018self,huang2018music} encode the relative distance between two elements in the sequence, rather than their absolute positions. The model learns to compute the relative distances between any two positions during training and looks up the corresponding embedding vectors in a pre-defined table to obtain the relative position embeddings. These embeddings are used to directly modify the attention matrix.
Position encoding has been verified to be effective in natural language processing and computer vision \cite{dufter2022position}.  However, in time series classification, the efficacy is still unclear.

The original absolute position encoding is proposed for language modeling, where high embedding dimensions like 512 or 1024 are usually used for position embedding of input with a length of 512 \cite{vaswani2017attention}. But, for time series tasks, embedding dimensions are relatively low, and the series might have a variety of lengths (ranging from very low to very high). In this paper, for the first time, we study the efficiency (i.e. how well resources are utilized) and the effectiveness (i.e. how well the encodings achieve their intended purpose) of existing absolute and relative position encodings for time series data. We then show that the existing absolute position encodings 
are ineffective with time series data. We introduce a novel time series-specific absolute position encoding method that takes into account the series embedding dimension and length. We show that our new absolute position encoding outperforms the existing absolute position encodings in time series classification tasks.

Additionally, since the existing relative position encodings have large memory overhead and they require a large number of parameters to be trained, in time series data it is very likely they overfit. We propose a novel computationally efficient implementation of relative position encoding to improve their generalisability for time series. We show that our new relative position encoding outperforms the existing relative position encodings in time series classification tasks.
We then propose a novel time series classification model based on the combination of our proposed absolute/relative position encodings named ConvTran to improve the position embedding of time series data. We further enriched the data embedding of time series using CNN rather than linear encoding. Our extensive experiments on 32 benchmark datasets show ConvTran is significantly more accurate than the previous state-of-the-art in deep learning models for time series classification (TSC). We believe our novel position encodings can boost the performance of other transformer-based TSC models.

\section{Related Work}
In this section, we briefly discuss the state-of-the-art multivariate time series classification (MTSC) algorithms, as well as CNN and attention-based models that have been applied to MTSC tasks.
We refer interested readers to the corresponding papers or the recent survey on deep learning for time series classification~\cite{foumani2023deep} for a more detailed description of these algorithms and models.

\subsection{State-of-the-art MTSC Algorithms}
Many MTSC algorithms have been proposed in recent years \cite{bagnall2018uea,ruiz2020great,fawaz2019deep}, where many of them are adapted from their univariate version.
A recent survey \cite{ruiz2020great} evaluated most of the existing MTSC algorithms on the UEA MTS archive, that consists of 26 equal-length time series datasets.
This benchmark includes a few deep learning as well as non-deep learning approaches.
This survey concluded that there are four main state of the art methods. 
These are ROCKET \cite{dempster2020rocket}, HIVE-COTE \cite{bagnall2020usage}, CIF \cite{middlehurst2020canonical} and Inception-Time \cite{fawaz2020inceptiontime}.

ROCKET \cite{dempster2020rocket} is a scalable TSC algorithm that uses 10,000 random convolution kernels to extract 2 features from each input time series, creating 20,000 features for each time series.
Then a linear model is used for classification, such as ridge or logistic regression.
Mini-ROCKET \cite{dempster2021minirocket} is an extension of ROCKET with some slight modifications to the feature extraction process.
It is significantly more scalable than ROCKET and uses only 10,000 features without compromising accuracy.
Multi-ROCKET \cite{tan2021multirocket} extends Mini-ROCKET by leveraging the first derivative of the series as well as extracting 4 features per kernel. 
It is significantly more accurate than both ROCKET and Mini-ROCKET on 128 univariate TSC tasks.
Note that neither Mini-ROCKET nor Multi-ROCKET has previously been benchmarked on the UEA MTS archive.
The adaptation for multivariate time series for ROCKET, Mini-ROCKET and Multi-ROCKET is done by randomly selecting different channels of the time series for each convolutional kernel.

The Canonical Interval Forest (CIF) \cite{middlehurst2020canonical} is an interval based classifier. 
It first extracts 25 features from random intervals of the time series and builds a time series forest with 500 trees.
It is an algorithm initially designed for univariate TSC and was adapted to multivariate TSC by expanding the random interval search space, where an interval is defined as a random dimension of the time series.

The Hierarchical Vote Collective of Transformation-based Ensembles (HIVE-COTE) is a meta ensemble for TSC. 
It forms its ensemble from classifiers of multiple domains.
Since its introduction in 2016, HIVE-COTE has gone through a few iterations. 
The version used in the MTSC benchmark \cite{ruiz2020great} comprised of 4 ensemble members -- Shapelet Transform Classifier (STC), Time Series Forest (TSF), Contractable Bag of Symbolic Fourier Approximation Symbols (CBOSS) and Random Interval Spectral Ensemble (RISE), each of them being the state of the art in their respective domains.
Since these algorithms were designed for univariate time series, the adaption for multivariate time series is not easy.
Hence, they were adapted for multivariate time series through ensembling over all the models built on each dimension independently. 
This means that they are computationally very expensive especially when the number of channels is large.
Recently, the latest HIVE-COTE version, HIVE-COTEv2.0 (HC2) was proposed \cite{middlehurst2021hive}. 
It is currently the most accurate classifier for both univariate and multivariate TSC tasks \cite{middlehurst2021hive}.
Despite being the most accurate on 26 benchmark MTSC datasets, that are relatively small, HC2 is not scalable to either large datasets with long time series or datasets with many channels.

\subsection{CNN Based Models}
CNNs are popular deep learning architectures for MTSC due to their ability to extract latent features from the time series data efficiently. 
Fully Convolutional Neural Network (FCN) and Residual Network (ResNet) were proposed in \cite{wang2017time} and evaluated in \cite{fawaz2019deep}. FCN is a simple convolutional network that does not contain any pooling layers in convolution blocks. 
The output from the last convolution block is averaged with a Global Average Pooling (GAP) layer and passed to a final softmax classifier. ResNet is one of the deepest architectures for MTSC (and TSC in general), containing three residual blocks followed by a GAP layer and a softmax classifier. It uses residual connections between blocks to reduce the vanishing gradient effect in deep learning models. 
ResNet was one of the most accurate deep learning TSC architectures on 85 univariate TSC datasets \cite{fawaz2019deep,bagnall2017great}. It was also proven to be an accurate deep learning model for MTSC \cite{fawaz2019deep,ruiz2020great}.


Inception-Time is the current state-of-the-art deep learning model for both univariate TSC and MTSC \cite{fawaz2020inceptiontime,ruiz2020great}.
Inception-Time is an ensemble of five randomly initialised inception network models that each consists of two blocks of inception modules. 
Each inception module first reduces the dimensionality of a multivariate time series using a bottleneck layer with length and stride of 1 while maintaining the same length. 
Then, 1D convolutions of different lengths are applied to the output of the bottleneck layer to extract patterns at different sizes.
A max pooling layer followed by a bottleneck layer are also applied to the original time series to increase the robustness of the model to small perturbations.
Residual connections are also used between each inception block to reduce the vanishing gradient effect. 
The output of the second inception block is passed to a GAP layer before feeding into a softmax classifier. 

Recently, Disjoint-CNN \cite{foumani2021disjoint} shows that factorization of 1D convolution kernels into disjoint temporal and spatial components yields accuracy improvements with almost no additional computational cost. Applying disjoint temporal convolution and then spatial convolution behaves similarly to the ``Inverted Bottleneck'' \cite{sandler2018mobilenetv2}. Like the Inverted Bottleneck, the temporal convolutions expand the number of input channels, and spatial convolutions later project the expanded hidden state back to the original size to capture the temporal and spatial interaction. 

\subsection{Attention Based Models}

Self-attention has been demonstrated to be effective in various natural language processing tasks due to its higher capacity and superior ability to capture long-term dependencies in text \cite{vaswani2017attention}. 
Recently, it has also been shown to be effective for time series classification tasks. Cross Attention Stabilized Fully Convolutional Neural Network (CA-SFCN) \cite{hao2020new} has applied the self-attention mechanism to leverage the long-term dependencies for the MTSC task. CA-SFCN combines FCN and two types of self-attention - temporal attention (TA) and variable attention (VA), which interact to capture both long-range temporal dependencies and  interactions between variables. With evidence that multi-headed attention dominates self-attention, many models try to adapt it to the MTSC domain. Gated Transformer Networks (GTN) \cite{liu2021gated}, similar to CA-SFCN, use two-tower multi-headed attention to capture discriminative information from the input series. They merge the output of two towers using a learnable matrix named gating. 

Inspired by the development of transformer-based self-supervised learning like BERT \cite{kostas2021bendr}, many models try to adopt the same structure for  time series classification \cite{kostas2021bendr, zerveas2021transformer}. BErt-inspired Neural Data Representations (BENDER) replace the word2vec encoder in BERT with the wav2vec to leverage the same structure for time series data. BENDER shows that if we have a massive amount of EEG data, the pre-trained model can be used effectively to model EEG sequences recorded with differing hardware. Similarly, Voice-to-Series with Transformer-based Attention (V2Sa) uses a large-scale pre-trained speech processing model for downstream problems like time series classification problems \cite{yang2021voice2series}.  
Recently, a Transformer-based Framework (TST) was also introduced to adopt vanilla transformers to the multivariate time series domain \cite{zerveas2021transformer}. TST uses only the encoder part of transformers and pre-train it with proportionally masked data in an unsupervised manner.


\section{Background}
This section provides a basic definition of self-attention and an overview of current position encoding models. Note that position encoding refers to the method that integrates position information, e.g., absolute or relative. Position embedding refers to a numerical vector associated with position encoding. 

\subsection{Problem Description and Notation}
Given a time series dataset $X$ with $n$ samples, $X=\left\{\mathbf{x_1},\mathbf{x_2},...,\mathbf{x_n}\right\}$, where $\mathbf{x_t} =\left\{x_1,x_2,...,x_L\right\}$ is a $d_x$-dimensional time series and $L$ is the length of time series, $\mathbf{x_t}\in \mathbb{R}^{L\times d_x}$, and the set of relevant response labels $Y=\left\{y_1,y_2,...,y_n \right\}$, $y_t\in\left\{1,...,c\right\}$ and c is the number of classes. The aim is to train a neural network classifier to map set $X$ to $Y$.

\subsection{Self-Attention}
The first attention mechanisms were proposed in the context of natural language processing \cite{luong2015effective}. While they still relied on a recurrent neural network at its core, Vaswani et al. \cite{vaswani2017attention} proposed a transformer model that relies on attention only. Transformers map a query and a set of key-value pairs to an output. More specifically, for an input series, $\mathbf{x_t} =\left\{x_1,x_2,...,x_L\right\}$, self-attention computes an output series $\mathbf{z_t} =\left\{z_1,z_2,...,z_L\right\}$ where $z_i\in \mathbb{R}^{d_z} $ and is computed as a weighted sum of input elements:
\vspace{-0.2cm}
\begin{equation} 
\label{e1}
\vspace{-0.2cm}
z_i=\sum_{j=1}^L \alpha_{i,j}(x_j W^V)
\end{equation}
Each coefficient weight $\alpha_{i,j}$ is calculated using softmax function:
\begin{equation}
\label{e2}
\alpha_{i,j}=\frac{exp(e_{ij})}{\sum_{k=1}^L exp(e_{ik})}
\end{equation}
where $e_{ij}$ is an attention weight from positions $j$ to $i$ and is computed using a scaled dot-product:
\begin{equation}
\label{e3}
e_{ij}=\frac{(x_i W^Q)(x_j W^K)^T}{\sqrt{d_z}}
\end{equation}
The projections $W^Q, W^K, W^V \in \mathbb{R}^{d_x \times d_z}$ are parameter matrices and are unique per layer. Instead of computing self-attention once, Multi-Head Attention (MHA) \cite{vaswani2017attention} does so multiple times in parallel, i.e., employing $h$ attention heads. A linear transformation is applied to the attention head outputs and concatenated into the standard dimensions. 

\subsection{Position Encoding}
The self-attention layer cannot preserve time series positional information in the transformer architecture since the transformer contains no recurrence and convolution. However, the local positional information, i.e., the ordering of time series, is essential. The practical approach in transformer-based methods involves using multiple encoding \cite{huang2020improve,wu2021rethinking,dufter2022position}, such as absolute or relative positional encoding, to enhance the temporal context of time-series inputs. 

\subsubsection{Absolute Position Encoding}
The original self-attention considers the absolute position \cite{vaswani2017attention}, and adds the absolute positional embedding $P =(p_1,..., p_L)$ to the input embedding $x$ as:
\begin{equation}
\label{e4}    
x_i = x_i+p_i
\end{equation}
where the position embedding $p_i \in \mathbb{R}^{d_{model}}$. There are several options for absolute positional encodings, including the fixed encodings by sine and cosine functions with different frequencies called $Vanilla APE$ and the learnable encodings through trainable parameters (we refer it as $Learn$ method) \cite{vaswani2017attention, devlin2018bert}. 

By using sine and cosine for fixed position encoding, the $d_{model}$-dimensional embeddings of $i_{th}$ time step position can be represented by the following equation: 
\begin{equation}
    p_i(2k)= \mathrm{sin}\: i\omega_k  \quad p_i(2k+1) =\mathrm{cos}\, i\omega_k \quad \omega_k=10000^{-2k/d_{model}}
    \label{eq:apeV}
\end{equation}
where $k$ is in the range of $[0,\frac{d_{model}}{2}]$, $d_{model}$ is the embedding dimension and $\omega_k$ is the frequency term. Variations in $\omega_k$ ensure that no positions $<$ $10^4$ are assigned similar embeddings.

\subsubsection{Relative Position Encoding} \label{sec:Back-RPE}
In addition to the absolute position embedding, recent studies in natural language processing and computer vision also consider the pairwise relationships between input elements, i.e., relative position \cite{shaw2018self,huang2018music}. This type of method encodes the relative distance between the input elements $x_i$ and $x_j$ into vectors $p_{i,j}^Q,p_{i,j}^K,p_{i,j}^V \in \mathbb{R}^{d_z}$. The encoding vectors are embedded into the self-attention module, which modifies Equation~\ref{e1} and Equation~\ref{e3} as 
\begin{equation}
\label{e5}
z_i=\sum_{j=1}^L \alpha_{i,j}(x_j W^V + p_{i,j}^V)
\vspace{-0.3cm}
\end{equation}
\begin{equation}
\label{e6}
e_{ij}=\frac{(x_i W^Q + p_{i,j}^Q)(x_j W^K+ p_{i,j}^K)^T}{\sqrt{d_z}}
\end{equation}
By doing so, the pairwise positional relation is trained during transformer training.

Shaw et al. \cite{shaw2018self} proposed the first relative position encoding for self-attention. Relative positional information is supplied to the model on two levels: values and keys.  First, relative positional information is included in the model as an additional component to the keys.  The softmax operation Equation~\ref{e3} remains unchanged from vanilla self-attention. Lastly, relative positional information is resupplied as a sub-component of the values matrix. Besides, the authors believe that relative position information is not useful beyond a certain distance, so they introduced a clip function to reduce the number of parameters. Encoding is formulated as follows to consider the distance between inputs $i$ and $j$ in computing their attention:
\begin{equation}
\label{e7}
e_{ij}=\frac{(x_i W^Q)(x_j W^K+ p_{clip(i-j,k)}^K)^T}{\sqrt{d_z}}
\vspace{-0.4cm}
\end{equation}
\begin{equation}
\label{e8}
z_i=\sum_{j=1}^L \alpha_{i,j}(x_j W^V + p_{clip(i-j,k)}^V)
\vspace{-0.1cm}
\end{equation}
\begin{equation}
\label{e9}
clip(x,k)= max (-k, min(k,x))
\end{equation}
Where $p^V$ and $p^K$ are the trainable weights of relative position encoding on values and keys, respectively. $P^V = (p_{-k}^V, ..., p_{k}^V )$ and $P^K = (p_{-k}^K, ..., p_{k}^K )$ where $p_{i}^V, p_{i}^K \in \mathbb{R}^{d_z}$. The scalar k is the maximum relative distance. 

However, this technique (Shaw) is not memory efficient.  As can be seen in Equation~\ref{e7}, it requires $O(L^2d)$ memory due to the additional relative position encoding. Huang et al. \cite{huang2018music} introduced a new method (in this paper it is called \textit{Vector} method) of computing relative positional encoding that reduces its intermediate memory requirement from $O(L^2d)$ to $O(Ld)$ using skewing operation \cite{huang2018music}. According to this paper, the authors also dropped the additional relative positional embedding corresponding to the value term and focused only on the key component. Encoding is formulated as follows: 
\begin{equation}
\label{e10}
e_{ij}=\frac{(x_i W^Q)(x_j W^K)^T + S^{rel}}{\sqrt{d_z}}
\vspace{-0.3cm}
\end{equation}
\begin{equation}
\label{e11}
S^{rel}=Skew(W^QP)
\end{equation}
Where $Skew$ procedure use padding, reshaping and slicing to reduce the memory requirement \cite{huang2018music}.
In Table \ref{tab1} we provided a summary of the parameter sizes, memory, and computation complexities of various position encoding methods (including our proposed ones in this paper) for comparison purposes.



\section{Position Encoding of Transformers for MTSC}

We design our position encoding methods to examine several aspects which are not well studied in prior transformers-based time series classification work (see the analysis in Sec \ref{sec-Ablation}). 

As a first step, we propose a new absolute position encoding method dedicated to time series data called \textbf{t}ime \textbf{A}bsolute \textbf{P}osition \textbf{E}ncoding (tAPE). tAPE incorporates the series length and input embedding dimension in absolute position encoding. We then introduce \textbf{e}fficient \textbf{R}elative \textbf{P}osition \textbf{E}mbedding (eRPE) to explore the independent encoding of positions from the input encodings. After that, to study the integration of eRPE into a transformer model, we compare different integration of position information to the attention matrix; finally, we provide an efficient implementation for our methods.

\subsection{Time Absolute Position Encoding (tAPE)}

\begin{figure}
  \centering
  \subfloat[]{%
    \includegraphics[trim=2.5cm 0cm 1.5cm 0cm, width=0.49\linewidth]{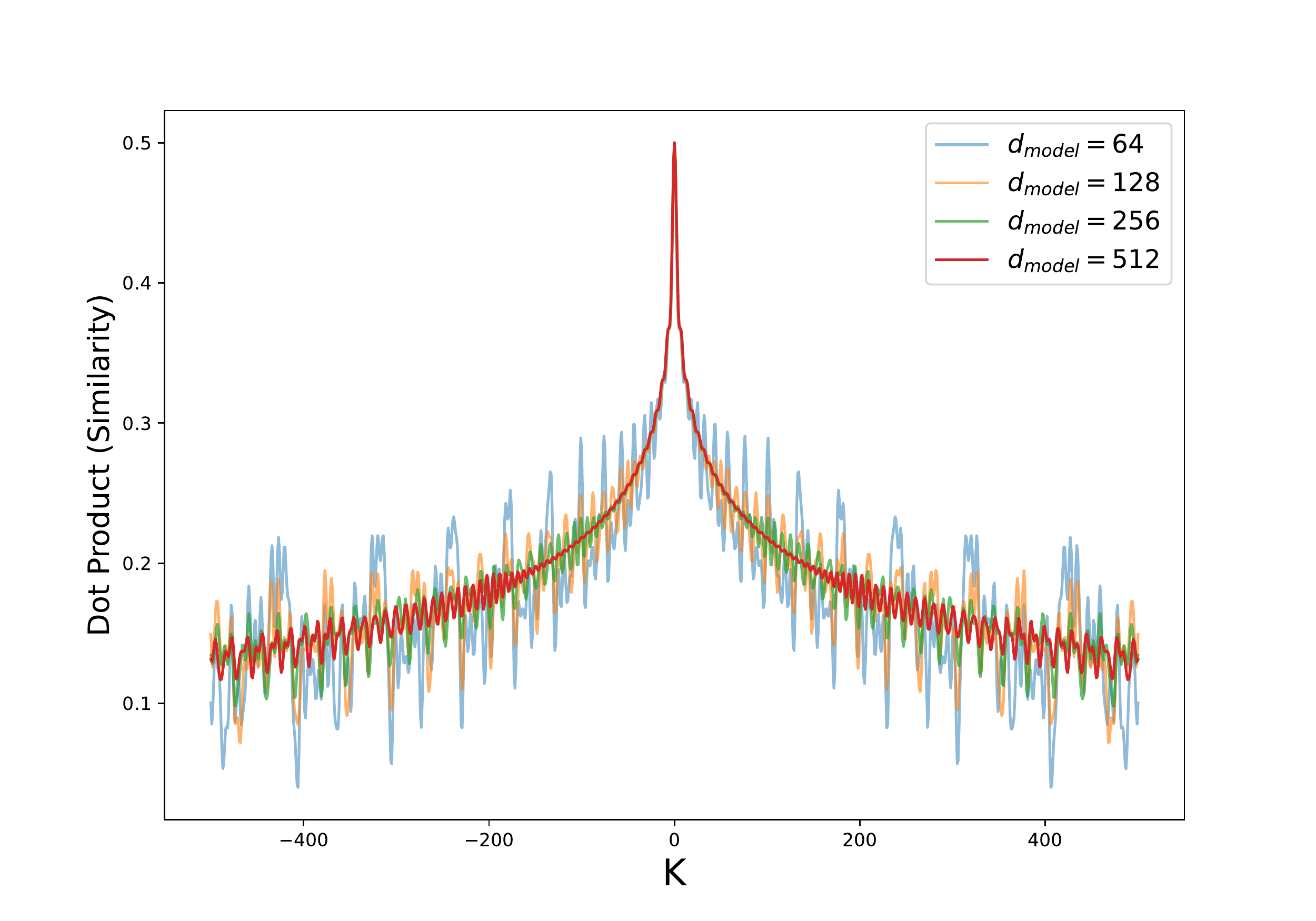}
    \label{fig:Sim-sin}%
  }\hfill%
  \subfloat[]{%
    \includegraphics[trim=1.5cm 0cm 2.5cm 0cm, width=0.49\linewidth]{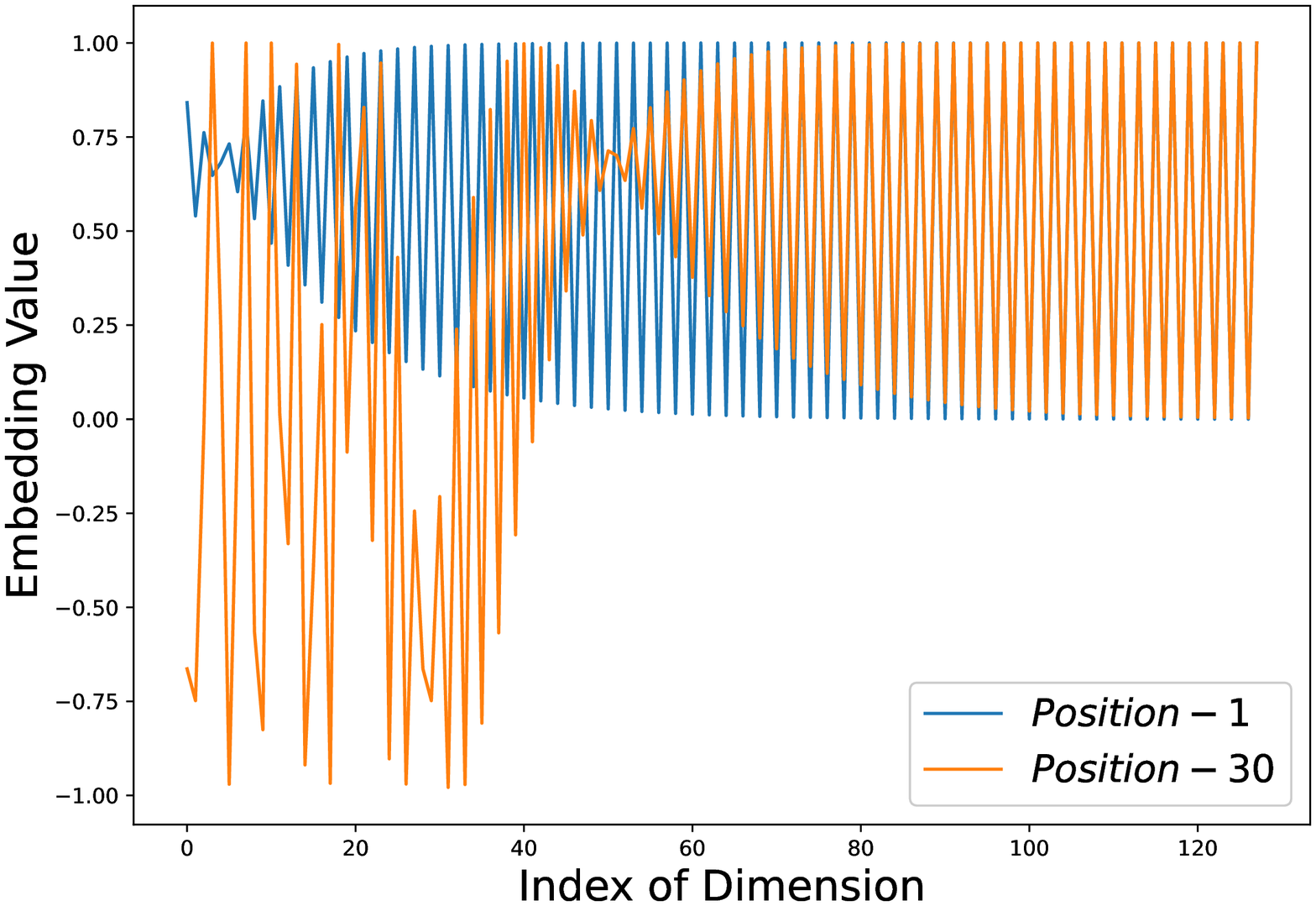}
    \label{fig:isotopic}%
  }
  \caption{Sinusoidal absolute position encoding. a) The dot product of two sinusoidal position embeddings whose distance is $K$ with various embedding dimensions.  b) 128 dimension sinusoidal positional encoding curves for positions 1 and 30 in a series of length 30.}
  \label{fig:Vanilla-embedding}
\end{figure}


Absolute position encoding was originally proposed for language modeling, where high embedding dimensions like 512 or 1024 are usually used for position embedding of input with a length of 512 \cite{vaswani2017attention}. Fig.\ref{fig:Sim-sin} shows the dot product between two sinusoidal positional embedding whose distance is $K$ using Equation~\ref{eq:apeV}  with various embedding dimensions. Clearly, higher embedding dimensions, such as 512 (red thick line), can better reflect the similarity between various positions. As shown in Fig.\ref{fig:Sim-sin} using 64 or 128 as embedding dimensions (thin blue and orange lines, respectively), the dot product does not always decrease as the distance between two positions increases. We call this the \textit{distance awareness property}, which disappears when lower embedding dimensions, such as 64, are used for position encoding. 

While high embedding dimensions show a desirable monotonous decrease trend when the distance between two positions increases (see red line in Fig.\ref{fig:Sim-sin}), they are not suitable for encoding time series datasets. The reason is that most time series datasets have relatively low input dimensionality (e.g., 28 out of 32 datasets have less than 64 input dimension), and higher embedding dimensions may yield inferior model throughput due to extra parameters (increasing the chances of overfitting the model). 

On the other hand, in low embedding dimensions, the similarity value between two random embedding vectors is high, making the embedding vectors very similar to each other. %
In other words, we cannot fully utilise the embedding vector space to differentiate between two positions. Fig. \ref{fig:isotopic} depicts the embedding vectors of the first and last position embedding for the embedding dimension equals 128 and length equals 30. In this figure, almost half of the embedding vectors are the same. This is called the \textit{anisotropic phenomenon} \cite{liang2021learning}. The anisotropic phenomenon makes the position encoding to be ineffective in low embedding dimensions as embedding vectors become similar to each other as it is shown in  Fig.\ref{fig:Sim-sin} (the blue line).

Hence, we require a position embedding for time series that has distance awareness  while simultaneously being isotropic. In order to incorporate distance awareness, we propose to use the time series length in Equation~\ref{eq:apeV}. 
In this equation, $\omega_k$ refers to the frequency of the sine and cosine functions from which the embedding vectors are generated. 
Without our modification, as series length $L$ increases the dot product of positions becomes ever less regular, resulting in a loss of distance awareness. By incorporating the length parameter in the frequency terms in both sine and cosine functions in Equation~\ref{eq:apeV}, the dot product remains smoother with a monotonous trend.  

As the embedding dimension $d_{model}$ value increases, it is more likely the vector embeddings are sampled from low-frequency sinusoidal functions, which results in the anisotropic phenomenon. To alleviate this, we incorporate the $d_{model}$ parameter into the frequency term in both sine and cosine functions in Equation~\ref{eq:apeV}. We propose a novel absolute position encoding for time series called tAPE in which $\omega_k^{new}$ takes into account the input embedding dimension and length as follows: 
\begin{alignat}{1}
\label{eq:tAPE}
    \omega_k=10000^{-2k/d_{model}} \nonumber\\
    \omega_k^{new} = \frac{\omega_k\times d_{model}}{L}
\end{alignat}
where $L$ is the series length and $d_{model}$ shows the embedding dimension. 

\begin{figure}
  \centering
  \subfloat[]{%
    \includegraphics[trim=2.5cm 0cm 1.5cm 0cm, width=0.45\linewidth]{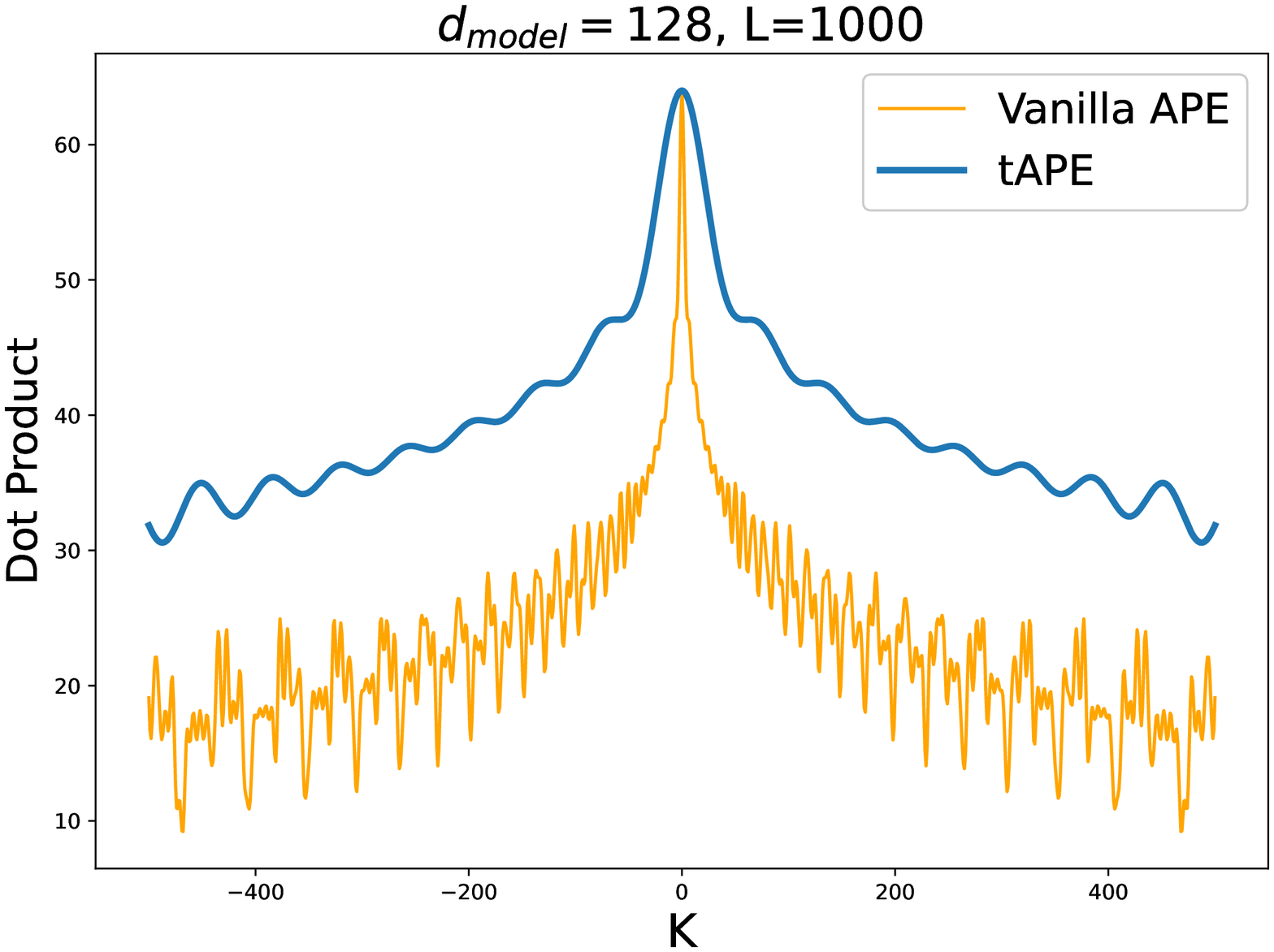}
    \label{fig:tAPE-a}%
  }\hfill%
  \subfloat[]{%
    \includegraphics[trim=1.5cm 0cm 2.5cm 0cm, width=0.45\linewidth]{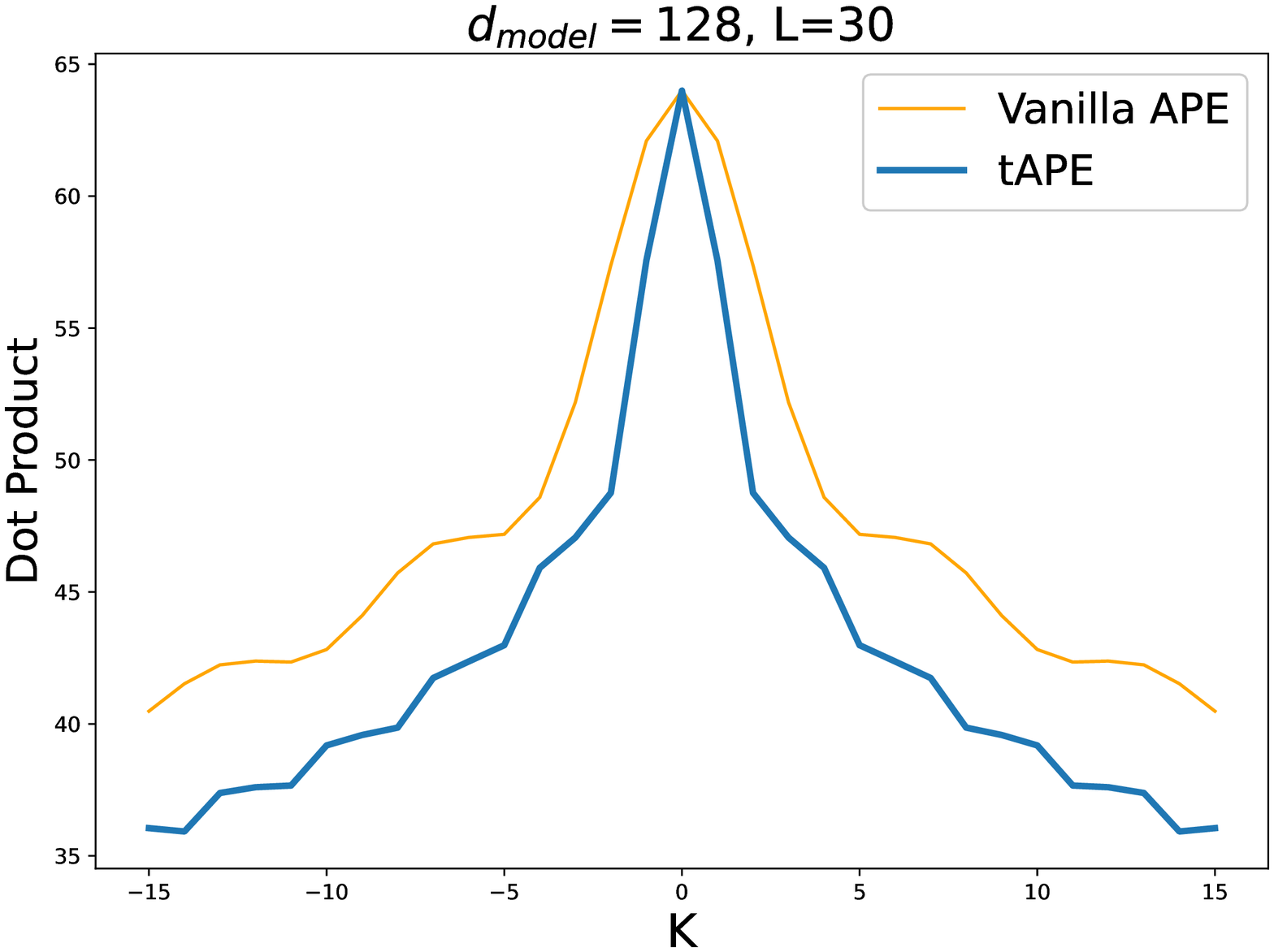}
    \label{fig:tAPE-b}%
  }
  \caption{Comparing dot product between two position whose distance is $K$ in a time series using tAPE and vanilla APE with $d_x=128$ dimension vector for series of length a) $L=1000$  b) $L=30$. }
  \label{fig:tAPE}
\end{figure}

Our new tAPE position encoding is compared with a vanilla sinusoidal position encoding to provide further illustration. Using $d_{model}=128$ dimension vector, Figs \ref{fig:tAPE}a-b show the dot product (similarity) of two positions with a distance of $K$ for series with of length $L=1000$ and $L=30$ respectively. As depicted in Fig \ref{fig:tAPE-a}, in vanilla APE, only the closest positions in the series have a monotonous decreasing trend, and approximately from a distance 50 onwards ($\lvert K\lvert>50$) on both sides, the decreasing similarity trend becomes less apparent as the distance between two positions in the time series increases. 
However, tAPE has a more stable decreasing trend and more steadily reflects the distance between two positions.  
Meanwhile, Fig \ref{fig:tAPE-b} shows the embedding vectors of tAPE are less similar to each other compared to vanilla APE. This is due to better utilising the embedding vector space to differentiate between two positions as we discussed earlier.

Note in Equation~\ref{eq:tAPE} our $\omega_k^{new}$ will obviously be equal to the  $\omega_k$ in vanilla APE when $d_{model}=L$ and the encodings of tAPE and vanilla APE will be the same. However, if $d_{model} \neq L$, tAPE will encode the positions in series more effectively than vanilla APE due to the two properties we discussed earlier. Fig \ref{fig:tAPE-a} shows a case in which $d_{model}<L$ and Fig \ref{fig:tAPE-b} shows a case in which $d_{model}>L$ and in both cases tAPE utilises embedding space to provide an isotropic encoding, while holding the distance awareness property. In other words, tAPE provides a balance between these two properties in its encodings. The superiority of tAPE compared to vanilla APE and learned APE on various length time series datasets is shown in the experimental results section.



\subsection{Efficient Relative Position Encoding (eRPE)}

There are multiple extensions of the abovementioned Section~\ref{sec:Back-RPE} relative position embeddings in machine translation and computer vision \cite{huang2020improve,wu2021rethinking,dufter2022position}. However, input embeddings are the basis for all previous methods of relative position encoding (adding or multiplying the position matrices to the query, key, and value matrices). In this study, we introduce an efficient model of relative position encoding independent of input embeddings.

\begin{figure}[ht]
    \centering
    \includegraphics[trim=2cm 1cm 6cm 1cm, width=0.7\textwidth]{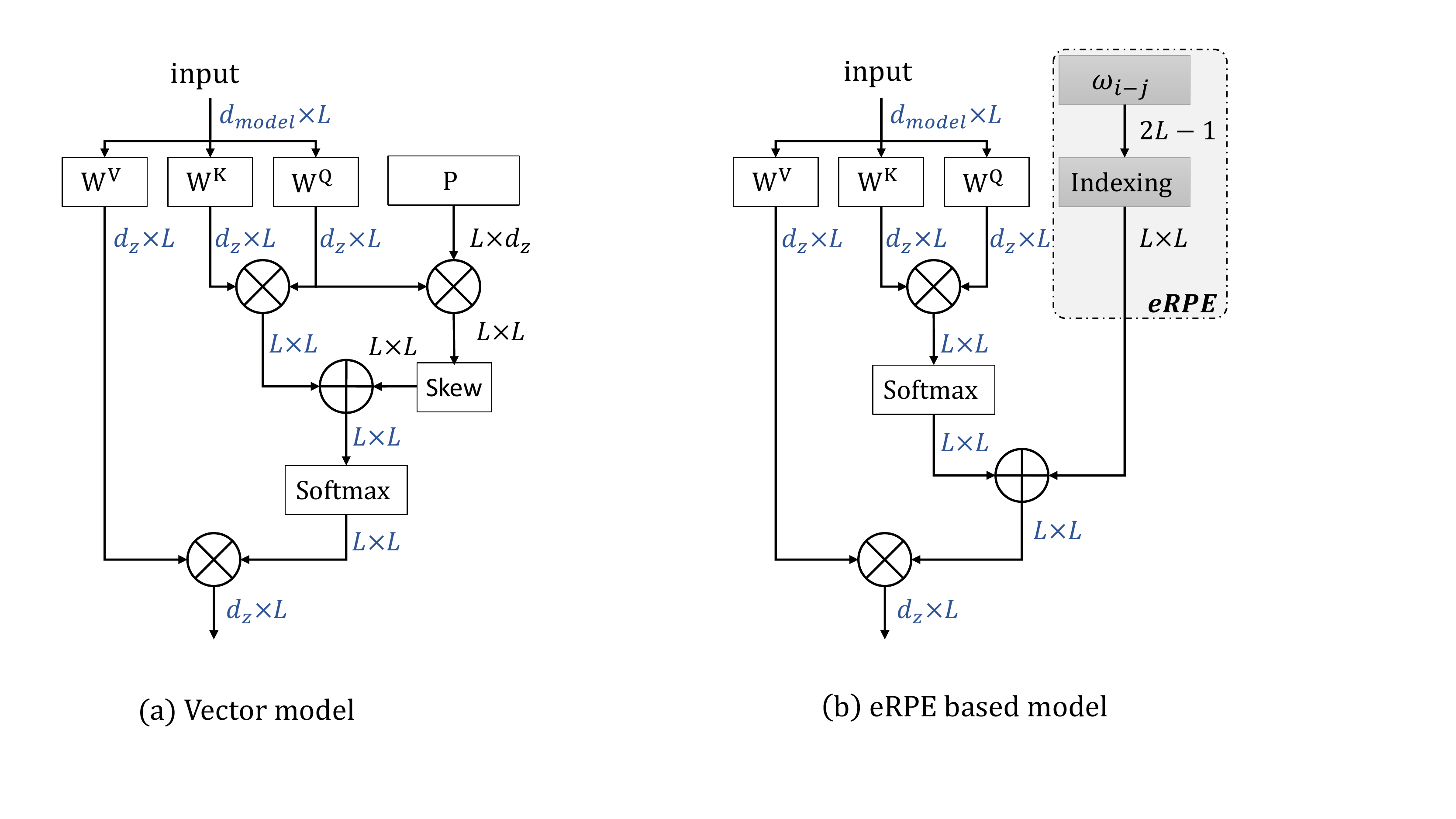}
    \caption{Self-attention modules with relative position encoding using scalar and vector parameters. Newly added parts are depicted in grey.}
    \label{fig:VecSca}
\end{figure}
In particular, we propose the following formulation: 
\begin{equation}
\alpha_{i}=\sum_{j\in L}\left(\underbrace{\frac{exp(e_{i,j})}{\sum_{k\in L}exp(e_{i,k})}}_{A_{i,j}}+w_{i-j}\right) x_j
\label{eq:at_rel}
\end{equation}
where $L$ is series length, $A_{i,j}$ is attention weight and $w_{i-j}$ is a learnable scalar (i.e., $w\in \mathbb{R}^{O(L)}$) and represent the relative position weight between positions $i$ and $j$. 

It is worth comparing the strengths and weaknesses of relative position encodings and attention to determine what properties are more desirable for relative position encoding of time series data. Firstly, the relative position embedding $w_{i-j}$ is an input-independent parameter with static values, whereas an attention weight $A_{i,j}$ is dynamically determined by the representation of the input series. In other words, attention adapts to input series via a weighting strategy (input-adaptive weighting \cite{vaswani2017attention}). Input-adaptive-weighting enables models to capture the complicated relationships between different time points, a property that we desire most when we want to extract high-level concepts in time series. This can be for instance the seasonality component in time series. However, when we have limited size data we are at a greater risk of overfitting when using attention.
 
Secondly, relative position embedding $w_{i-j}$ takes into account the relative shift between positions $i$ and $j$ and not their values. This is similar to translation equivalence property of convolution, which has been shown to enhance generalization \cite{dai2021coatnet}. We propose to consider the notation of $w_{i-j}$ as a scalar rather than a vector to enable the translation equivalency without blowing up the number of parameters. In addition, the scalar representation of $w$ provides the benefit that the value of $w_{i-j}$ for all $(i, j)$ can be subsumed within the pairwise dot-product attention function, resulting in minimal additional computation (see subsection \ref{Efficient_Implementation}). We call our proposed efficient relative position encoding as eRPE.

Theoretically, there are many possibilities for integrating relative position information into the attention matrix, but we empirically found that attention models perform better when we add the relative position to the model after applying the softmax to the attention matrix as shown in Equation~\ref{eq:at_rel}. We presume this is because the position values will be sharper without the softmax. And sharper position embeddings seems to be beneficial in TSC task as it emphasizes more on informative relative positions for classification compared to existing models in which softmax is applied to relative position embeddings.

\subsubsection{Efficient Implementation: Indexing} \label{Efficient_Implementation}
To implement the efficient version of eRFE in Equation~\ref{eq:at_rel} for input time series with a length of $L$,
for each head, we create a trainable parameter $w$ of size $2L-1$, as the maximum distance is $2L-1$. Then for two position indices $i$ and $j$, the corresponding relative scalar is $w_{i-j+L}$ where indexes start from 1 instead of 0 (1-base index). 
 Accordingly, we need to index $L^2$ elements from $2L-1$ vector. 

On GPU, a more efficient way to index is to use \texttt{gather}, which only requires memory access. At inference time, indexing the $L^2$ elements from $2L-1$ vector can be pre-computed and cached to increase the processing speed further. 
As shown in Table \ref{tab1}, our proposed eRPE is more efficient in terms of both memory and time complexities compared to the existing relative position encoding methods in the literature. 

\begin{table}[]
\centering
\setlength{\tabcolsep}{1pt}
\setlength\extrarowheight{0pt}
\caption{Comparing the parameter sizes, memory, and computation complexities of various position encoding methods. In our implementation $d_z$ is equal to $d_{model}$}.
\begin{tabular}{ccccc}
\hline 
\multicolumn{2}{c}{\textbf{Method}} & \textbf{Parameter} & \textbf{Memory} & \textbf{Complexity} \\ \hline
\multicolumn{1}{c}{} & tAPE & None & $Ld_{model}$ & $Ld_{model}$ \\ \cline{2-5} 
\multicolumn{1}{c}{\multirow{2}{*}{\textbf{Absolute}}} & Vanilla APE \cite{vaswani2017attention} & None & $Ld_{model}$ & $Ld_{model}$ \\
\cline{2-5} 
\multicolumn{1}{c}{} & Learn \cite{devlin2018bert} & $Ld_{model}$ & $Ld_{model}$ & $Ld_{model}$ \\ \hline
\multicolumn{1}{c}{\multirow{3}{*}{\textbf{Relative}}} & Shaw \cite{shaw2018self} & $(2L-1)d_z$ & $L^2d_z+L^2$ & $L^2d_z$ \\ \cline{2-5} 
\multicolumn{1}{c}{} & Vector \cite{huang2018music} & $Ld_z$ & $Ld_z+L^2$ & $L^2d_z$ \\ \cline{2-5} 
\multicolumn{1}{c}{} & eRPE & $2L-1$ & $L+L^2$ & $L^2$ \\ \hline
\label{tab1}
\end{tabular}
\end{table}

\subsection{ConvTran}

\begin{figure*}
    \centering
    \includegraphics[trim=1cm 11.5cm 4cm 2cm, width=0.98\textwidth]{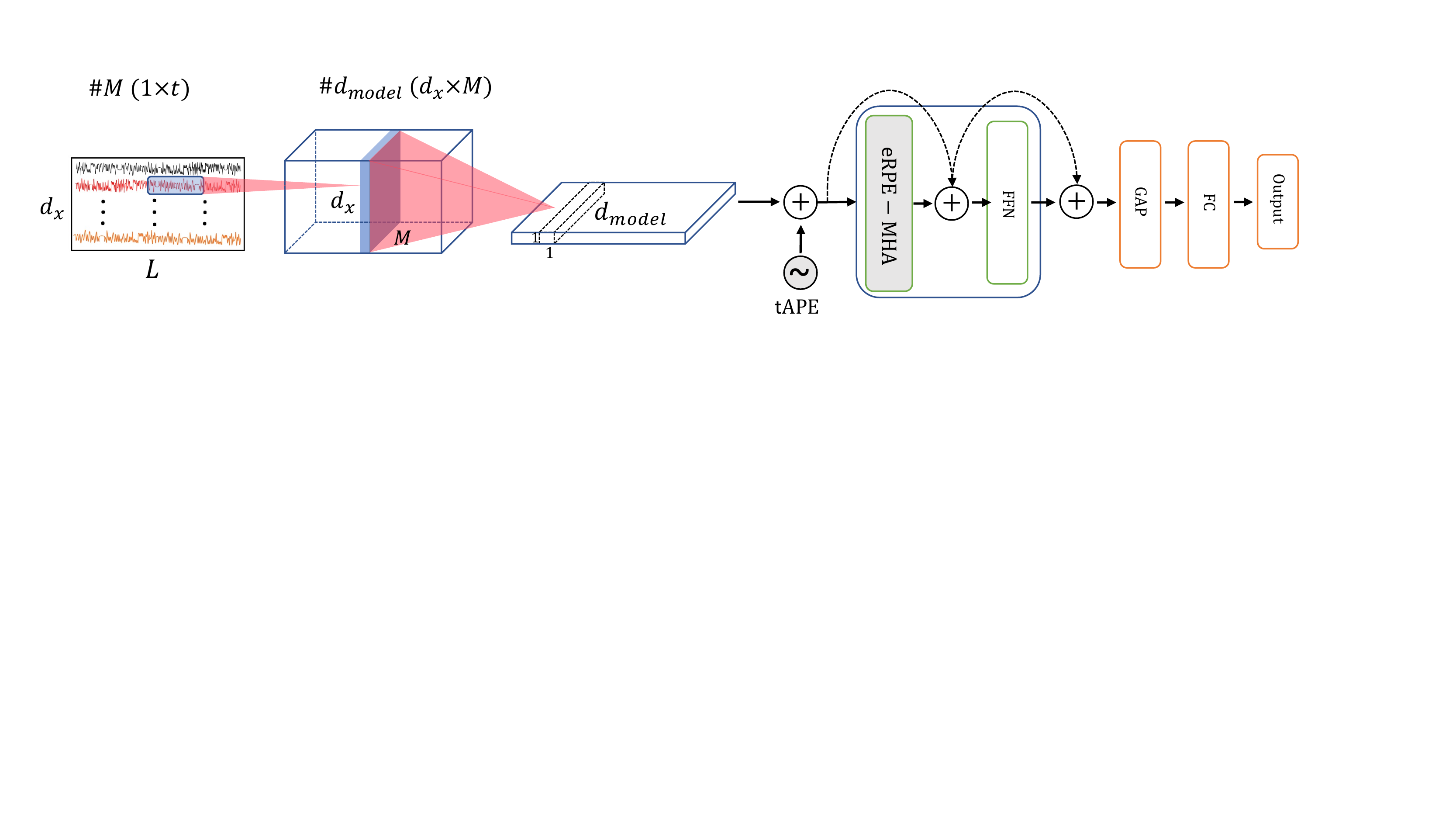}
    \caption{Overall Architecture of the ConvTran Model}
    \label{fig:BlockDiagram}
\end{figure*}

Now we look at how we can utilize our new position encodings method to build a time series classification network. According to the earlier discussion, global attention has a quadratic complexity w.r.t. the series length. This means that if we directly apply the proposed attention in Equation~\ref{eq:at_rel} to the raw time series, the computation will be excessively slow for long time series. 
Hence, we first use convolutions to reduce the series length and then apply our proposed position encodings once the feature map has been reduced to a less computationally intense size. See Fig. \ref{fig:BlockDiagram} where convolution blocks comes as a first component proceeded by attention blocks.

Another benefit of using convolutions is that convolutions operations are very well-suited to capture local patterns. By using convolutions as the first component in our architecture we can capture any discriminative local information that exists in raw time series.

As Shown in Fig. \ref{fig:BlockDiagram}, as the first step in the convolution layers, $M$ temporal filters are applied to the input data. In this step, the model extracts temporal patterns in the input series. Next, the output of temporal filtering is convolved with $d_{model}$ spatial $d_x\times M$ shape filters to capture the correlations between variables in multivariate time series and construct $d_{model}$ size input embeddings.
Such disjoint temporal and spatial convolution is similar to ``Inverted Bottleneck'' in \cite{sandler2018mobilenetv2}. It first expands the number of input channels and then squeezes them. A key reason for this choice is that the Feed Forward Network (FFN) in transformers \cite{vaswani2017attention} also expands on the input size and later projects the expanded hidden state back to the original size to capture the spatial interactions.


Before feeding the input embedding to the transformer block, we add the tAPE-generated position embedding to the input embedding vector so that the model can capture the temporal order of the time series. The size of the embedding vector is $d_model$, which is the same as the input embedding. Inside the multi-head attention, the inputs with the $L\times d_{model}$ dimension 
are first converted to $L\times d_{z}\times3$ shape using a linear layer to get the \texttt{qkv} matrix in which $d_z$ indicates the model dimension and defined by the user. Each of the three matrices of shape $L\times d_z$ represents the Query (q), Key (k) and Value (v) matrices. These q, k, and v matrices are reshaped to $h\times L\times d_z/h$  to represent the $h$ attention heads. Each of these attention heads can be responsible for capturing different patterns in time series. For instance, one attention head can attend to the non-noisy data, another head can attend to the seasonal component and another to the trend. Once we have the $q$, $k$, and $v$ matrices, we finally perform the attention operation inside the Multi-Head attention block using Equation~\ref{eq:at_rel}.

According to Equation~\ref{eq:at_rel} the eRPE with the same shape of $L\times L$ is also added to the attention output. We consider the notation of $w_{i-j}$ as a scalar (i.e., $w\in R^{O(L)}$) to enable the global convolution kernel without increasing the number of parameters. 
The relative position embedding enables the model to learn not only the order of time points, but also the relative position of pairs of time points, which can capture richer information than other position embedding strategies.

The FFN, is a multi-layer perceptron block consisting of two linear layers and Gaussian Error Linear Units (GELUs) as an activation function. The outputs from the FFN block are again added to the inputs (via skip connection) to get the final output from the transformer block. Finally, just before the fully connected layer, max-pooling and global average pooling (GAP) are applied to the output of the last layer's ELU activation function, which gives a more translation-equivalence model.

\section{Experimental Results} \label{Sec:Results}
In this section, we evaluate the performance of our ConvTran model on the UEA time series repository \cite{bagnall2018uea} and two large multivariate time series datasets and compare it with the state-of-the-art models. 
All of our experiments were conducted using the PyTorch framework in Python on a computing system consisting of a single Nvidia A5000 GPU with 24GB of memory and an Intel(R) Core(TM) i9-10900K CPU.
To promote reproducibility, we have provided our source code and more experimental results online \footnote{\url{https://github.com/Navidfoumani/ConvTran}}.

We have divided our experiments into four parts. First, we present an ablation study on various position encodings. Then, we demonstrate that our ConvTran model outperforms existing CNN and transformer-based models. Next, we compare the performance of ConvTran with four state-of-the-art MTSC algorithms (including both deep learning and non-deep learning categories) identified in~\cite{ruiz2020great,middlehurst2021hive}. We report the results provided on the archive website\footnote{\url{https://timeseriesclassification.com/HC2.php}} for HiveCote2, CIF, ROCKET, and Inception-Time on 26 out of 30 UEA datasets only in Section~\ref{Sec:Benchmark}. Finally, we evaluate the efficiency and effectiveness of ConvTran by comparing it with the current state-of-the-art model, ROCKET.

\subsection{Datasets}
\begin{itemize}
\item[] \textbf{UEA Repository}
The archive consists of 30 real-world multivariate time series data from a wide range of applications such as Human Activity Recognition, Motion classification, and ECG/EEG classification \cite{bagnall2018uea}. 
The number of dimensions ranges from two dimensions to 1345 dimensions.
The length of the time series ranges from 8 to 17,984. 
The datasets also have a train size ranging from 12 to 25000.

\item[] \textbf{Ford Challenge}
This dataset is obtained from the Kaggle challenge website \footnote{\url{https://www.kaggle.com/c/stayalert}}. It includes measurements from total of 600 real-time driving sessions where each driving session takes 2 minutes and sampled with 100ms rate. Also, the trials are samples from 100 drivers of both genders, and of different ages. The training data file consists of 604,329 data points each belongs to one of 500 trials. The test file contains 120,840 data points belonging to 100 trials. While each data point comes with a label in {0,1} and also contains 8 physiological, 12 environmental, and 10 vehicular features that are acquired while driving. 

\item[] \textbf{Actitracker human Activity Recognition}
This dataset describes six daily activities which are collected in a controlled laboratory environment.
The activities include ``Walking'', ``Jogging'', ``Stairs'', ``Sitting'', ``Standing'', and ``Lying Down'' which are recorded from 36 users collected using a cell phone in their pocket. Data has 2,980,765 samples with 3 dimensions, subject-wise split into train and test sets, and a sampling rate of 20Hz~\cite{lockhart2011design}.
\end{itemize}

\subsection{Evaluation Procedure}
We use the classification accuracy as the overall metric to compare different models. Then we rank each model based on its classification accuracy per dataset. The most accurate model is assigned a rank of 1 and the worse performing model is assigned the highest rank. The average ranking is taken in case of ties.  Then the average rank for each model is computed across all datasets in the repository.

This gives a direct general assessment of all the models: the lowest rank corresponds to the method that is the most accurate on average. 
The average ranking for each model is presented in the form of critical difference diagram \cite{demvsar2006statistical}, where models in the same clique (the black bar in the diagram) are not statistically significant. 
For the statistical test, we used the Wilcoxon signed-rank test with Holm correction as the post hoc test to the Friedman test \cite{demvsar2006statistical}. 

\subsection{Parameter Setting}
Adam optimization is used simultaneously with an early stopping method based on validation loss. We use the default setting for other models.
We set the default value for the number of temporal and spatial filters to 64 and set the length of the temporal filters to 8. The width of the spatial convolutions are set equal to the input dimensions \cite{foumani2021disjoint}.

Similar to TST, the transformers based model for MTSC \cite{zerveas2021transformer}, and default transformers block \cite{vaswani2017attention}, we use 8 heads to capture the varieties of attention from input series. The dimension of transformers encoding is set to $d_{model} = d_z = 64$ and FFN in transformers block expands the input size by 4x and later projects the 4x-wide hidden state back to the original size.


\subsection{Ablation Study on Position Encoding} \label{sec-Ablation}
In this section, firstly we compare our proposed tAPE with the exisiting absolute position encodings. Secondly, we compare our proposed eRPE with the existing relative position encoding methods. As a final step, we combined tAPE and eRPE into a single framework and campare it with all possible combinations of absolute and relative position encodings.

\begin{figure}
  \centering
  \subfloat[]{%
    \includegraphics[trim=2cm 1cm 0cm 2.5cm, width=0.48\linewidth]{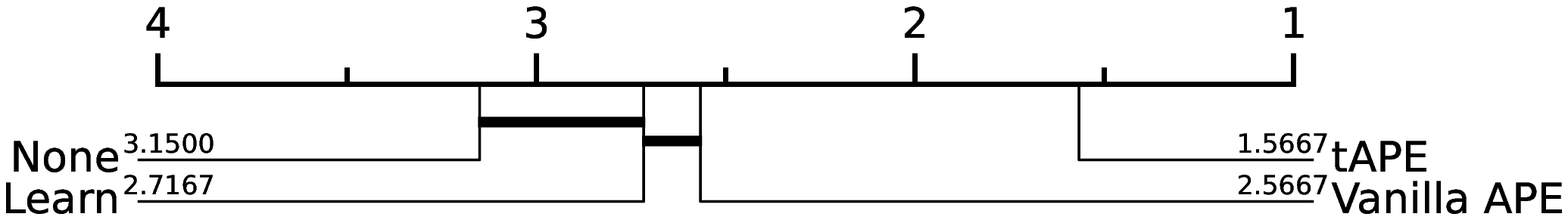}
    \label{fig:APE-CD}%
  }\hfill%
  \subfloat[]{%
    \includegraphics[trim=1.2cm 1cm 2.5cm 1cm, width=0.48\linewidth]{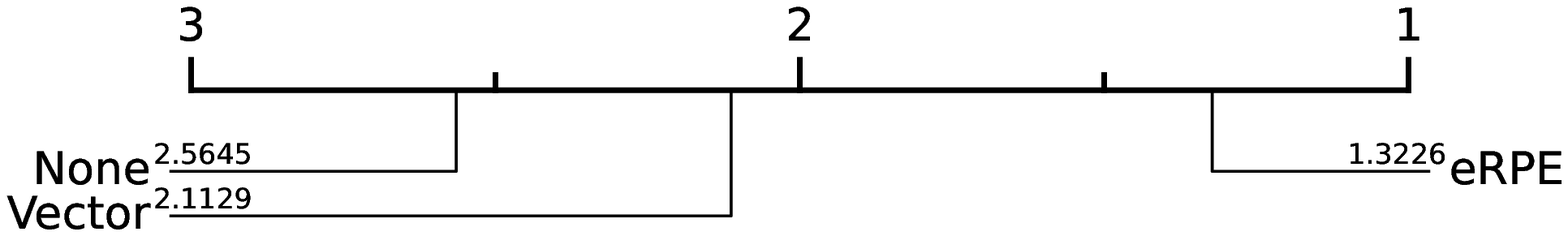}
    \label{fig:RPE-CD}%
  }
  \caption{Critical difference diagram of various position encoding over thirty datasets for the UEA MTSC archive based on average accuracies: a) Various absolute position encodings, b) Various relative position encodings. The lowest rank corresponds to the method that is the most accurate on average.}
  \label{fig:APE-RPE}
\end{figure}

For this ablation study we run a single-layer transformer five times on all 30 UEA benchmark datasets for classification. Fig.\ref{fig:APE-CD} illustrates the critical difference diagram of a single-layer transformer with different absolute position encodings. Note in critical difference diagram methods grouped by a black line are not significantly different from each other.  
In Fig.\ref{fig:APE-RPE}, \textit{None} is the model without any position encoding, \textit{Learn} is the model with learning absolute position encoding parameters \cite{devlin2018bert}, \textit{Vanilla APE} is the vanilla sinusoidal function-based encoding \cite{vaswani2017attention}, \textit{Vector} is the vector-based implementation of input-dependent relative position embedding  \cite{huang2018music}, and our proposed models showed as \textit{tAPE} and \textit{eRPE}. 

As depicted in Fig.\ref{fig:APE-CD}, tAPE has the highest rank in terms of accuracy and is significantly better than other absolute position encodings due to effectively utilising embedding space to provide an isotropic encoding while holding the distance awareness property. As expected, the model without position encoding has the least accurate results, highlighting the importance of absolute position encoding in time series classification. The vanilla APE also improves overall performance despite not being significantly accurate than Learn APE since it has fewer parameters. 

\begin{figure}
     \centering
     \includegraphics[width=0.8\columnwidth]{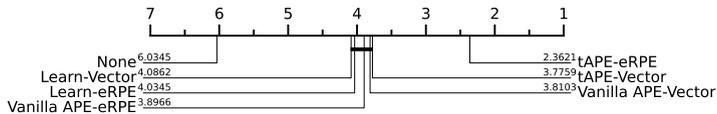}
     \caption{The average rank of various combination of absolute and relative position encodings.}
     \vspace*{-1pt}
     \label{fig:Combined-CD}
\end{figure}

Fig.\ref{fig:RPE-CD} shows the critical difference diagram of a single-layer transformer with different relative position encodings. As shown in this figure, eRPE has the highest rank and is significantly better than other encodings in terms of accuracy as it has less number of parameters which is less likely to overfit. It is not surprising that the model without position encoding has the least accurate results, highlighting the importance of relative position encoding and the translation equality property in time series classification. The input-dependent Vector encoding also improves overall performance and is significantly better than None model.
Fig.\ref{fig:Combined-CD} shows the critical difference diagram for the various combinations of absolute and relative position encodings. As depicted in this figure, the combination of our proposed tAPE and eRPE is significantly more accurate than all other combinations. This shows the high potential of our encoding methods to incorporate position information into transformers. The combination of Learn and Vector has the least accurate results, most likely due to the high number of parameters.  

\subsection{Comparing with State-of-the-Art Deep Learning Models}
We compare our ConvTran with the following convolution-based and transformer-based models for MTSC:
\begin{itemize}
\item[] \textbf{FCN}: Fully Convolutional Neural network is one of the most accurate deep neural networks for MTSC \cite{fawaz2019deep} reported in the literature. 
\item[] \textbf{ResNet}: Residual Network is also one of most accurate deep neural networks for both univariate TSC and MTSC\cite{fawaz2019deep} reported in the literature.
\item[] \textbf{Disjoint-CNN}: One of the accurate and lightweight CNN-based models that factorize convolution kernels into disjoint temporal and spatial convolutions \cite{foumani2021disjoint}.
\item[] \textbf{Inception-Time}: The most accurate deep learning univariate TSC and MTSC algorithm to date. \cite{fawaz2020inceptiontime,ruiz2020great}.
\item[] \textbf{TST}: A transformer-based model for MTSC \cite{zerveas2021transformer}.
\end{itemize}

\begin{figure}
     \centering
     \includegraphics[width=0.8\columnwidth]{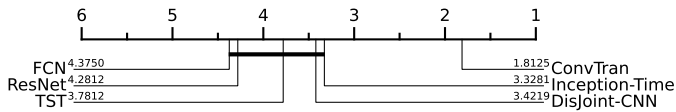}
     \caption{The average rank of ConvTran against all deep learning based methods on all 32 MTS datasets.}
     \vspace*{-1pt}
     \label{fig:CD-Deep}
\end{figure}

Fig. \ref{fig:CD-Deep} shows the average rank of ConvTran on 32 MTS datasets againts all convolutional-based and/or transformer-based methods.
This figure shows that on average, ConvTran has the lowest average rank and is more accurate than all other methods. 
It is important to observe that ConvTran is significantly more accurate than its predecessors, i.e., a convolution based model, Disjoint-CNN as well as the transformer based model, TST.
This indicates the effectiveness of adding tAPE and eRPE to transformers. Table \ref{tab:results} presents the classification accuracy of each method on all 32 datasets and the highest accuracy for each dataset is highlighted in bold. In this table datasets are sorted based on the number of training samples per class. Considering Fig. \ref{fig:CD-Deep} and Table \ref{tab:results} we can conclude that ConvTran is the most accurate TSC method on average on all 32 benchmark datasets and particularly has superior performance in datasets in which there are enough data to train (i.e., the number of training samples per class is more than 100) and wins on all 12 datasets except one.

\begin{table}
\centering
\setlength{\tabcolsep}{1.5 pt}
\setlength\extrarowheight{0.3 pt}
\caption{Average accuracy of six deep learning based models over 32 multivariate time series datasets. Datasets are sorted based on the number of training samples per-class. 
The highest accuracy for each dataset is highlighted in bold.}
\label{tab:results}
\begin{tabular}{|l|c|c|c|c|c|c|c|c|c|}
\hline
DataSets & Avg Train & ConvTran & TST & IT & Disjoint-CNN & FCN & ResNet \\ \hline \hline
Ford & 17300 & \textbf{0.7805} & 0.7655 & 0.7628 & 0.7422 & 0.6353 & 0.687 \\ \hline
HAR & 8400 & \textbf{0.9098} & 0.8831 & 0.8775 & 0.8807 & 0.8445 & 0.8711 \\ \hline
FaceDetection & 2945 & \textbf{0.6722} & 0.6542 & 0.5885 & 0.5665 & 0.5037 & 0.5948 \\ \hline
Insectwingbeat & 2500 & \textbf{0.7132} & 0.6748 & 0.6956 & 0.6308 & 0.6004 & 0.65 \\ \hline
PenDigits & 750 & \textbf{0.9871} & 0.9694 & 0.9797 & 0.9708 & 0.9857 & 0.9771 \\ \hline
ArabicDigits & 660 & \textbf{0.9945} & 0.9749 & 0.9872 & 0.9859 & 0.9836 & 0.9832 \\ \hline
LSST & 176 & \textbf{0.6156} & 0.2846 & 0.4456 & 0.5559 & 0.5616 & 0.5725 \\ \hline
FingerMovement & 158 & 0.56 & \textbf{0.58} & 0.56 & 0.54 & 0.53 & 0.54 \\ \hline
MotorImagery & 139 & \textbf{0.56} & 0.48 & 0.53 & 0.49 & 0.55 & 0.52 \\ \hline
SelfRegSCP1 & 134 & \textbf{0.918} & 0.86 & 0.8634 & 0.8839 & 0.7816 & 0.8362 \\ \hline
Heartbeat & 102 & \textbf{0.7853} & 0.6975 & 0.6248 & 0.717 & 0.678 & 0.7268 \\ \hline
SelfRegSCP2 & 100 & \textbf{0.5833} & 0.5333 & 0.4722 & 0.5166 & 0.4667 & 0.5 \\ \hline 
PhonemeSpectra & 85 & \textbf{0.3062} & 0.089 & 0.1586 & 0.2821 & 0.1599 & 0.1596 \\ \hline
CharacterTraject & 72 & 0.9922 & 0.9825 & 0.9881 & \textbf{0.9945} & 0.9868 & \textbf{0.9945} \\ \hline
EthanolConcen & 66 & \textbf{0.3612} & 0.151 & 0.3489 & 0.2775 & 0.3232 & 0.3155 \\ \hline
HandMovement & 40 & 0.4054 & \textbf{0.5405} & 0.3783 & \textbf{0.5405} & 0.2973 & 0.2838 \\ \hline
PEMS-SF & 39 & 0.8284 & 0.7572 & \textbf{0.8901} & \textbf{0.8901} & 0.8324 & 0.7399 \\ \hline
RacketSports & 38 & 0.8618 & \textbf{0.8815} & 0.8223 & 0.8355 & 0.8223 & 0.8223 \\ \hline
Epilepsy & 35 & 0.9855 & 0.9492 & \textbf{0.9928} & 0.8898 & \textbf{0.9928} & \textbf{0.9928} \\ \hline
JapaneseVowels & 30 & \textbf{0.9891} & 0.9837 & 0.9702 & 0.9756 & 0.973 & 0.9135 \\ \hline
NATOPS & 30 & 0.9444 & \textbf{0.95} & 0.9166 & 0.9277 & 0.8778 & 0.8944 \\ \hline
EigenWorms & 26 & \textbf{0.5934} & 0.4503 & 0.5267 & \textbf{0.5934} & 0.4198 & 0.4198 \\ \hline
UWaveGesture & 15 & 0.8906 & 0.8906 & \textbf{0.9093} & 0.8906 & 0.85 & 0.85 \\ \hline
Libras & 12 & \textbf{0.9277} & 0.8222 & 0.8722 & 0.8577 & 0.85 & 0.8389 \\ \hline
ArticularyWord & 11 & 0.9833 & 0.9833 & \textbf{0.9866} & \textbf{0.9866} & 0.98 & 0.98 \\ \hline
BasicMotions & 10 & \textbf{1} & 0.975 & \textbf{1} & \textbf{1} & \textbf{1} & \textbf{1} \\ \hline
DuckDuckGeese & 10 & \textbf{0.62} & 0.5 & 0.36 & 0.5 & 0.36 & 0.24 \\ \hline
Cricket & 9 & \textbf{1} & \textbf{1} & 0.9861 & 0.9772 & 0.9306 & 0.9722 \\ \hline
Handwriting & 6 & \textbf{0.3752} & 0.2752 & 0.3011 & 0.2372 & 0.376 & 0.18 \\ \hline
ERing & 6 & \textbf{0.9629} & 0.9296 & 0.9296 & 0.9111 & 0.9037 & 0.9296 \\ \hline
AtrialFibrillation & 5 & \textbf{0.4} & 0.2 & 0.2 & \textbf{0.4} & 0.3333 & 0.3333 \\ \hline
StandWalkJump & 4 & 0.3333 & 0.3333 & \textbf{0.4} & 0.3333 & \textbf{0.4} & \textbf{0.4} \\ \hline
\end{tabular}
\end{table}

\subsection{Benchmark against State-of-the-Art Models} \label{Sec:Benchmark}
Given the experiments on the 32 datasets show that our ConvTran model has the best performance compared to all the other convolution and transformers based models, we now proceed to benchmark it against the state-of-the-art MTSC models, i.e., both deep learning and non-deep learning models.
We compare HC2, CIF and ROCKET models on only 26 out of 32 MTSC benchmarking datasets \cite{ruiz2020great} because the other six datasets are either large in terms of training sample or have varied series lengths that make it almost impossible to run HC2 on them. 
For having detailed insights into the ConvTran performance we provide a pair-wise comparison between our proposed model and each of these models.

\begin{figure*}
  \centering
  \subfloat[]{%
    \includegraphics[trim=0 0 0 3, clip,width=0.44\linewidth]{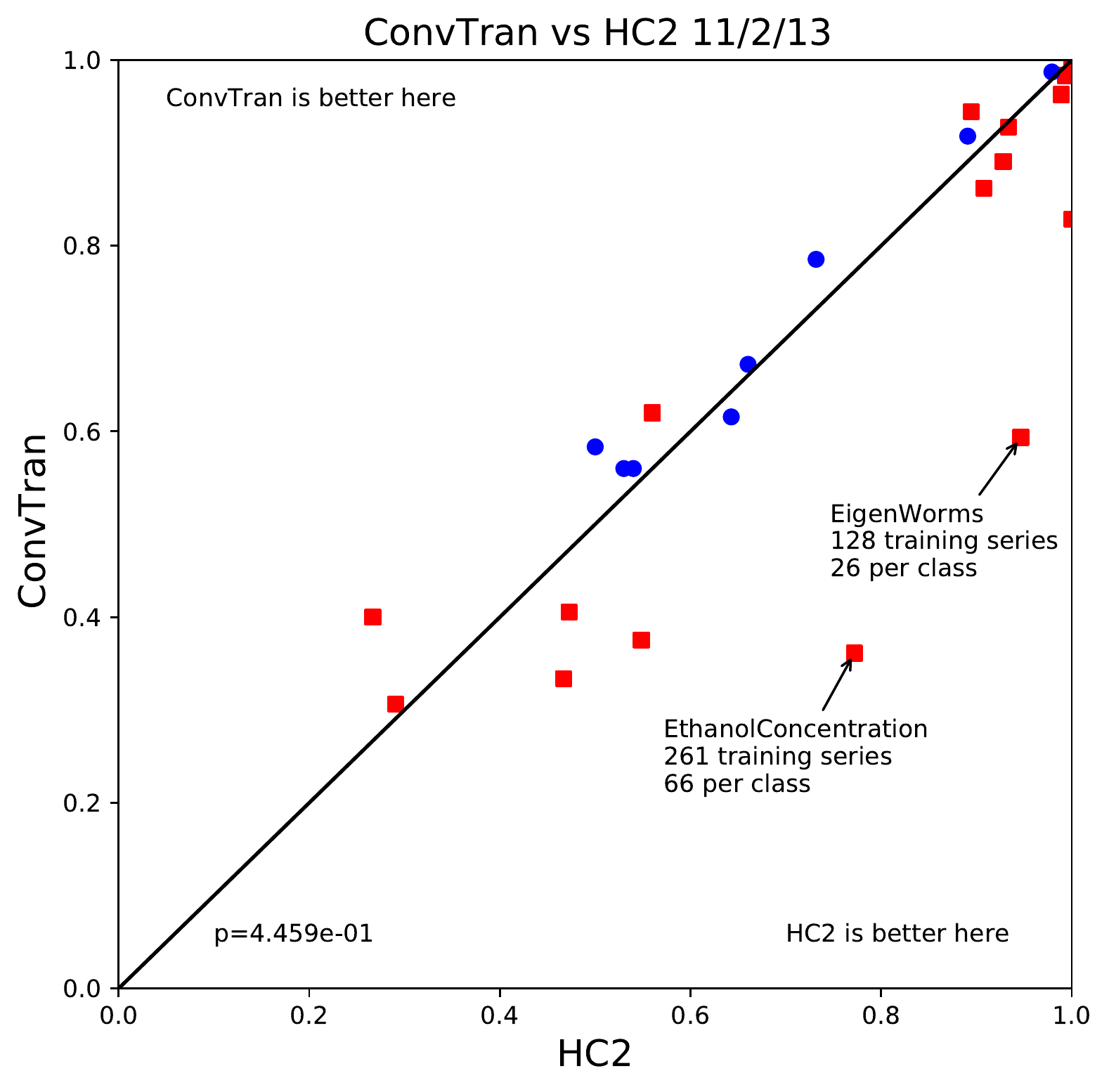}
    \label{fig:vs hc2}%
  }\hfill%
  \subfloat[]{%
    \includegraphics[trim=0 0 0 3, clip,width=0.44\linewidth]{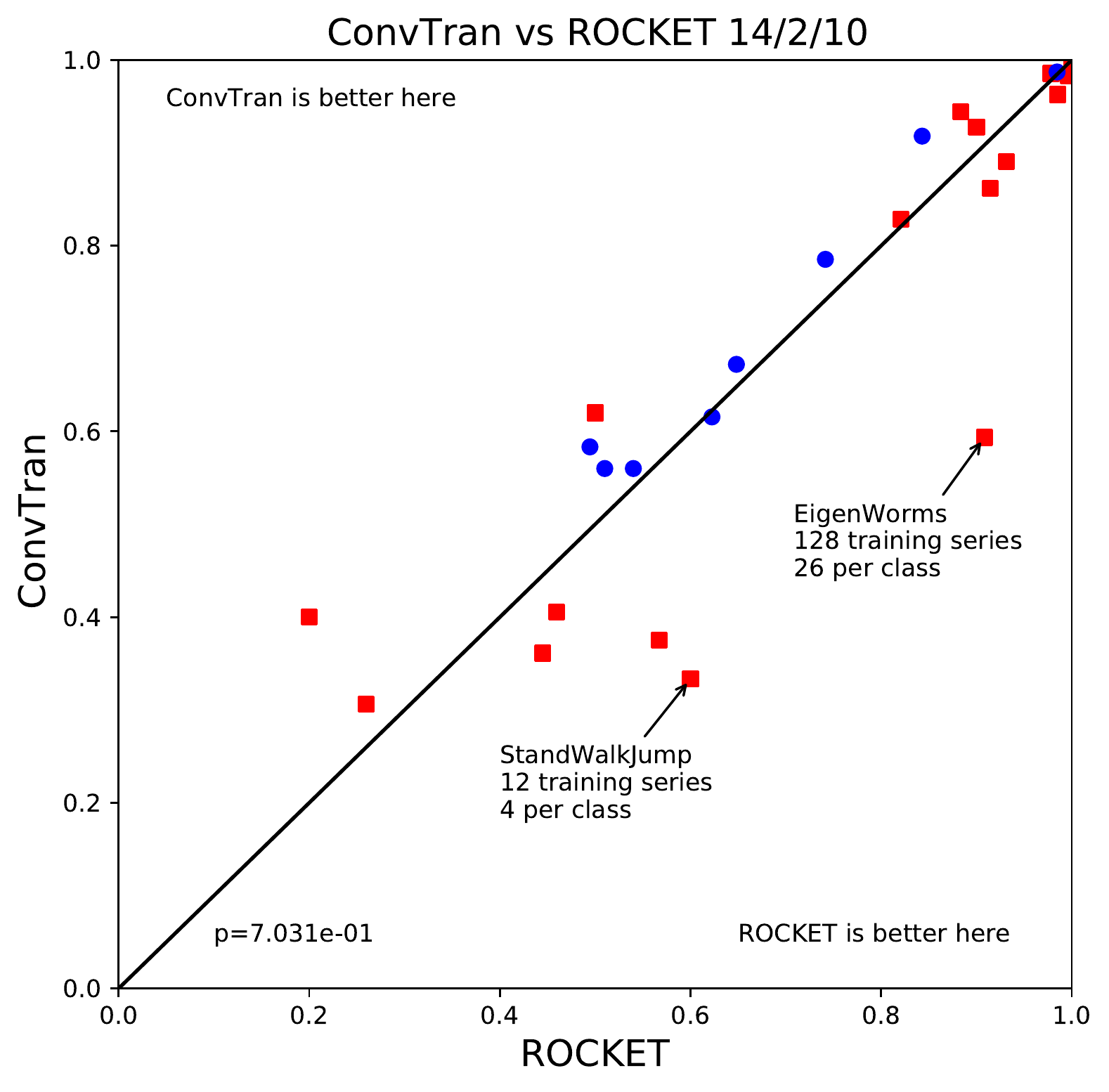}
    \label{fig:vs rocket}%
  }\hfill%
  \subfloat[]{%
    \includegraphics[trim=0 0 0 4, clip,width=0.44\linewidth]{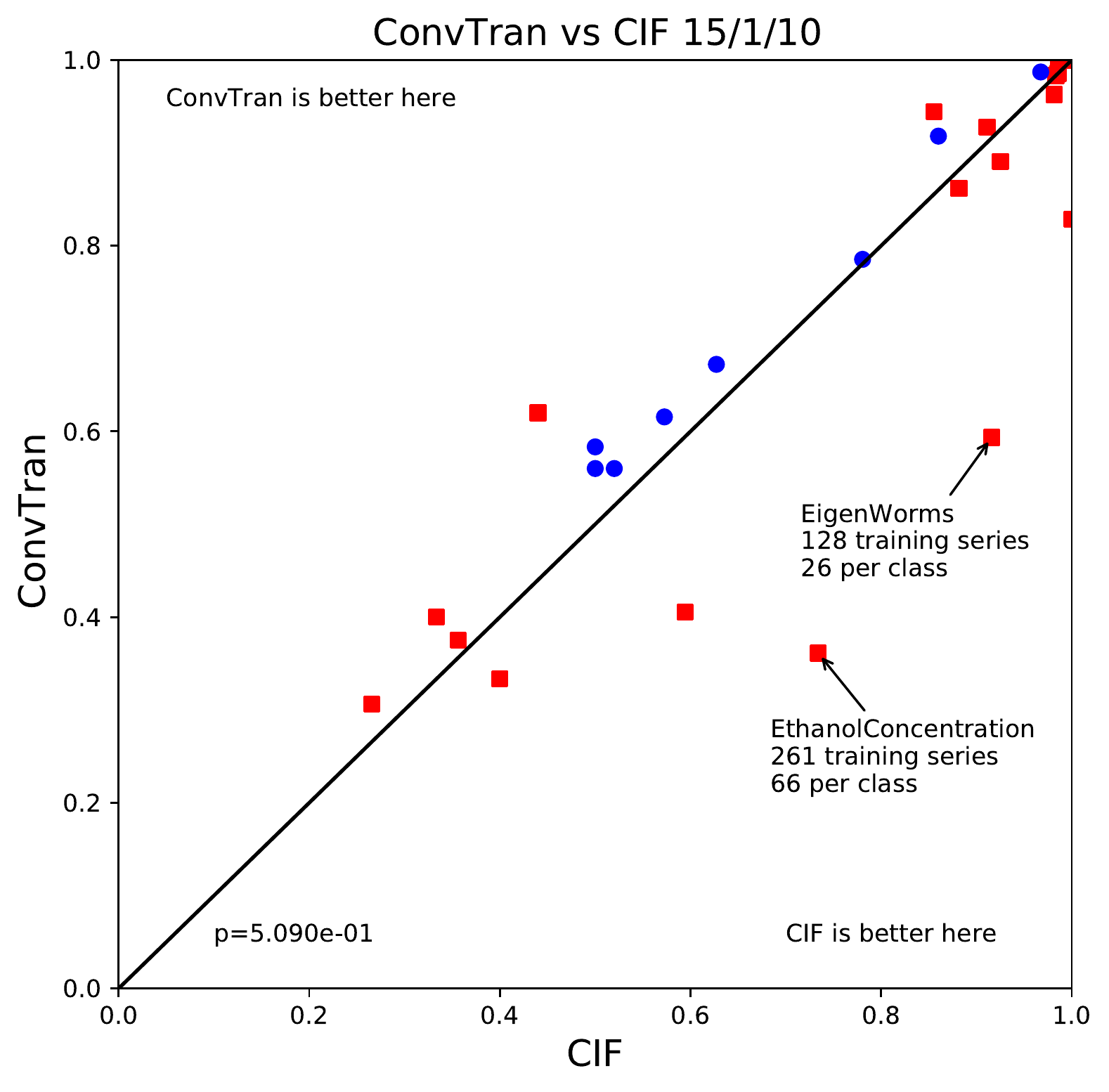}
    \label{fig:vs cif}%
  }\hfill%
  \subfloat[]{%
    \includegraphics[trim=0 0 0 4, clip,width=0.44\linewidth]{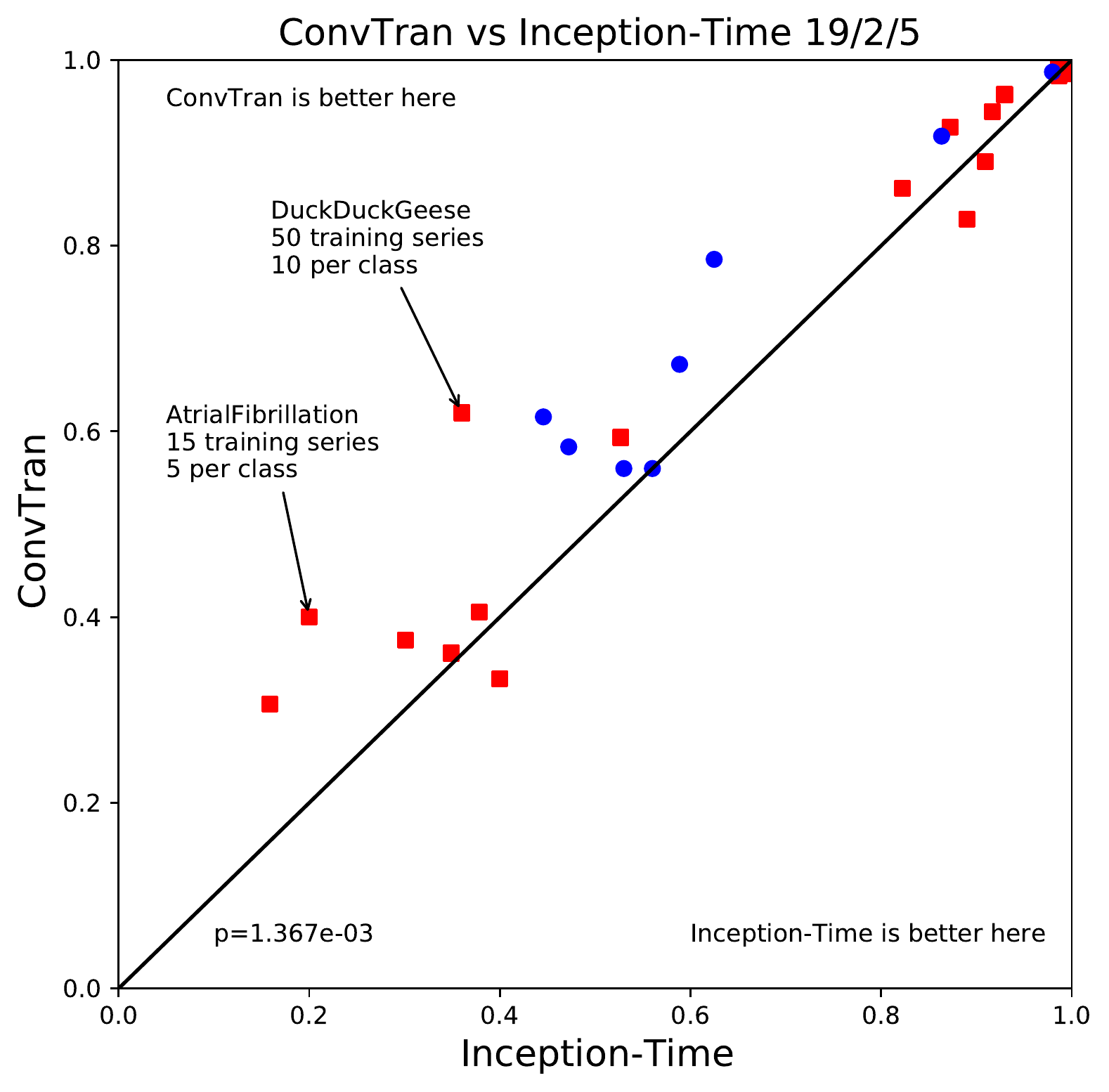}
    \label{fig:vs inception}%
  }%
  \caption{Pairwise comparison of ConvTran with the state of the art models: (a) HC2, (b) ROCKET, (c) CIF (d) and Inception-Time. The datasets with 100 training samples per class or more are marked with a blue circle, while the others are marked with a red square. The three values at the top of each figure show the number of win/draw/loss from left to right}
  \label{fig:pairwise}
\end{figure*}

As shown in Fig.~\ref{fig:pairwise} our proposed model mostly outperforms HC2, ROCKET, CIF, and Inception-Time on the datasets with 100 or more training samples per class (marked with a blue circle). However, state-of-the-art models outperform ConvTran on datasets with few training instances 
such as \texttt{EigenWorms} with 
26 train sample per-class. Indeed, as shown in Table \ref{tab:results}, all CNN based models fail to perform competitively on the \texttt{EigenWorms} dataset. Note that ConvTran is the most accurate among all CNNs on this dataset. This is due to the limitation of CNN-based models, which cannot capture long-term dependencies in the high length time series. Adding a transformer improves the performance, but it still requires more training samples to perform as well as other models.

It is also interesting to observe from Figs. \ref{fig:vs hc2} and \ref{fig:vs cif} that HC2 and CIF perform better than ConvTran on the \texttt{EthanolConcentration} dataset. Considering that this dataset is based on spectra of water-and-ethanol, hence interval and shapelet-based approaches which are also components of HC2 perform better.
On the other hand, ROCKET has a few wins compared to ConvTran (Fig \ref{fig:vs rocket}). Most of these datasets where ROCKET performs better, such as the \texttt{StandWalkjump} dataset have a small number of time series instances per class. 
For instance, \texttt{StandWalkjump} has 3 classes with 12 training instances, which is 4 time series per class. This is insufficient to train large number of parameters in deep learning models such as ConvTran to achieve better performance. Note, as mentioned, these results are for 26 datasets only, excluding six datasets for which we could not run HC2 (which has high computational complexity and is limited to be applied on variable-length time series). Among excluded datasets, 4 of them are large datasets from which ConvTran could have benefited. Considering this, ConvTran still achieves competetive performance compared to SOTA deep and non-deep models.

\subsection{ConvTran vs ROCKET Efficiency and Effectiveness}

To provide further insight into the efficiency of our model on datasets of varying sizes, we conducted additional experiments on the largest UEA dataset \textit{InsectWingBeat} with 25,000 series for training. We compare the training time and test accuracy of our proposed ConvTran and ROCKET on random subsets of 5,000, 10,000, 15,000, 20,000, and 25,000 training samples.
\begin{figure}
     \includegraphics[trim=1cm 0.5cm 1cm 1cm, width=0.95\columnwidth]{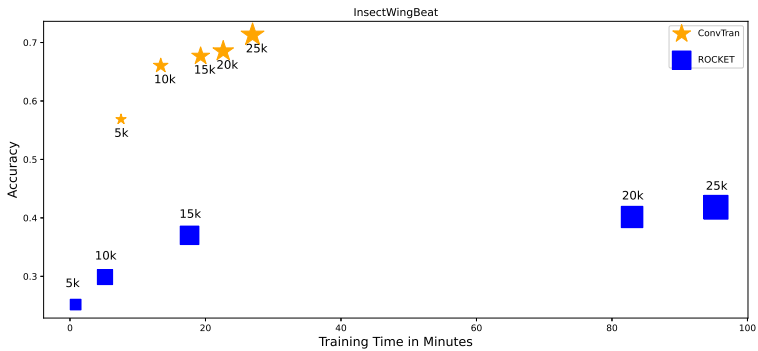}
     \caption{ Comparison of runtime and accuracy between ConvTran and ROCKET on UEA largest dataset InsectWingBeat with 25,000 training samples. The figure shows the runtime of the two models on datasets with different sizes, and their corresponding classification accuracy.}
     \label{fig:Runtime}
\end{figure}

The results depicted in Figure \ref{fig:Runtime} demonstrate that ROCKET has faster training time than ConvTran on smaller datasets, specifically on the 5k and 10k datasets while achieving similar training time to ConvTran on the 15k set. 
However, our deep learning-based model, ConvTran, demonstrates faster training times with increasing data quantity, as expected.  Additionally, we also observed from the figure that ConvTran is consistently more accurate than ROCKET on this dataset. We refer interested readers to Appendix~\ref{Appendix:A} for a more comprehensive exploration of the empirical evaluation of efficiency and effectiveness on all datasets. Notably, ConvTran demonstrates faster inference time compared to ROCKET across all datasets. It is important to note that all the ConvTran experiments are performed on GPUs, whereas ROCKET experiments are performed on a CPU (please refer to Section~\ref{Sec:Results} for computing system details).

\section{Conclusion}
This paper studies the importance of position encoding for time series for the first time and reviews existing absolute and relative position encoding methods in time series classification. Based on the limitations of the current position encodings for time series, we proposed two novel absolute and relative position encodings sepecifically for time series called tAPE and eRPE, respectively. We then integrated our two proposed position encodings into a transformer block and combine them with a convolution layer and presented a novel deep-learning framework for multivariate time series classification (ConvTran). Extensive experiments show that ConvTran benefits from the position information, achieving state-of-the-art performance on Multivariate time series classification in deep learning literature. In future, we will study the effectiveness of our new transformer block in other transformer-based TSC models and other down stream tasks such as anomaly detection.

\section{Declarations}

\textbf{Conflict of interest statement:} The authors have no competing interests to declare that are relevant to the content of this article.

\bibliography{bibliography}

\begin{appendices}

\clearpage 
\section{}

\subsection{Empirical Evaluation of Efficiency and Effectiveness}
\label{Appendix:A}
The results presented in Table~\ref{tab:Conv_ROCKET} demonstrate that ConvTran outperforms ROCKET in terms of both train time and test accuracy on larger datasets with more than 10k samples. 
However, ROCKET has a better train time on smaller datasets. Nevertheless, even on small datasets, ConvTran achieves acceptable accuracy within a reasonable train time. It is worth noting that the performance of ConvTran improves as the dataset size increases, indicating that our model is suitable for scaling to larger datasets.

\begin{table}
\centering
\setlength{\tabcolsep}{1.5 pt}
\setlength\extrarowheight{0.5 pt}
\caption{Comparison of runtime and accuracy between ConvTran and ROCKET on 32 datasets of varying sizes. To facilitate easy identification, superior performance in both accuracy and runtime is highlighted in bold in the table. For a detailed comparison, the runtimes are shown in seconds.}
\label{tab:Conv_ROCKET}
\begin{tabular}{|l|l|ccc|ccc|}
\hline
\multirow{2}{*}{\textbf{Datasets}} & \multirow{2}{*}{\textbf{Train size}} & \multicolumn{3}{c|}{\textbf{ROCKET}} & \multicolumn{3}{c|}{\textbf{ConvTran}} \\ 
 &  & \multicolumn{1}{c|}{Accuracy} & \multicolumn{1}{c|}{Train time} & Test time & \multicolumn{1}{c|}{Accuracy} & \multicolumn{1}{c|}{Train time} & Test time \\ \hline \hline
HAR & 41546 & \multicolumn{1}{c|}{0.8293} & \multicolumn{1}{c|}{5366.34} & 11.51 & \multicolumn{1}{c|}{\textbf{0.9098}} & \multicolumn{1}{c|}{\textbf{2367.77}} & \textbf{1.82} \\ \hline
Ford & 28839 & \multicolumn{1}{c|}{0.6051} & \multicolumn{1}{c|}{6863.81} & 11.91 & \multicolumn{1}{c|}{\textbf{0.7805}} & \multicolumn{1}{c|}{\textbf{1619.42}} & \textbf{0.95} \\ \hline
InsectWingbeat & 25000 & \multicolumn{1}{c|}{0.4182} & \multicolumn{1}{c|}{5721.04} & 41.5 & \multicolumn{1}{c|}{\textbf{0.7132}} & \multicolumn{1}{c|}{\textbf{1617.82}} & \textbf{5.47} \\ \hline
PenDigits & 7494 & \multicolumn{1}{c|}{0.984} & \multicolumn{1}{c|}{\textbf{65.26}} & 0.99 & \multicolumn{1}{c|}{\textbf{0.9871}} & \multicolumn{1}{c|}{401.1} & \textbf{0.59} \\ \hline
ArabicDigits & 6599 & \multicolumn{1}{c|}{0.9932} & \multicolumn{1}{c|}{\textbf{75.59}} & 10.38 & \multicolumn{1}{c|}{\textbf{0.9945}} & \multicolumn{1}{c|}{376.7} & \textbf{0.37} \\ \hline
FaceDetection & 5890 & \multicolumn{1}{c|}{0.5624} & \multicolumn{1}{c|}{\textbf{53.23}} & 11.99 & \multicolumn{1}{c|}{\textbf{0.6722}} & \multicolumn{1}{c|}{413.39} & \textbf{0.83} \\ \hline
PhonemeSpectra & 3315 & \multicolumn{1}{c|}{0.1894} & \multicolumn{1}{c|}{\textbf{42}} & 37.22 & \multicolumn{1}{c|}{\textbf{0.3062}} & \multicolumn{1}{c|}{202.27} & \textbf{0.89} \\ \hline
LSST & 2459 & \multicolumn{1}{c|}{0.5251} & \multicolumn{1}{c|}{\textbf{5.84}} & 3.52 & \multicolumn{1}{c|}{\textbf{0.6156}} & \multicolumn{1}{c|}{148.07} & \textbf{0.48} \\ \hline
CharacterTrajec & 1422 & \multicolumn{1}{c|}{0.9916} & \multicolumn{1}{c|}{\textbf{8.4}} & 7.72 & \multicolumn{1}{c|}{\textbf{0.9922}} & \multicolumn{1}{c|}{89.61} & \textbf{0.28} \\ \hline
FingerMovement & 316 & \multicolumn{1}{c|}{0.55} & \multicolumn{1}{c|}{\textbf{0.96}} & 0.35 & \multicolumn{1}{c|}{\textbf{0.56}} & \multicolumn{1}{c|}{21.33} & \textbf{0.02} \\ \hline
MotorImagery & 278 & \multicolumn{1}{c|}{\textbf{0.56}} & \multicolumn{1}{c|}{\textbf{45.39}} & 16.26 & \multicolumn{1}{c|}{\textbf{0.56}} & \multicolumn{1}{c|}{386} & \textbf{0.81} \\ \hline
ArticularyWord & 275 & \multicolumn{1}{c|}{\textbf{0.9933}} & \multicolumn{1}{c|}{\textbf{2.09}} & 2.19 & \multicolumn{1}{c|}{0.9833} & \multicolumn{1}{c|}{19.76} & \textbf{0.08} \\ \hline
JapaneseVowels & 270 & \multicolumn{1}{c|}{0.9568} & \multicolumn{1}{c|}{\textbf{0.57}} & 0.67 & \multicolumn{1}{c|}{\textbf{0.9891}} & \multicolumn{1}{c|}{20.6} & \textbf{0.13} \\ \hline
SelfRegSCP1 & 268 & \multicolumn{1}{c|}{0.8601} & \multicolumn{1}{c|}{\textbf{10.2}} & 11.25 & \multicolumn{1}{c|}{\textbf{0.918}} & \multicolumn{1}{c|}{45.54} & \textbf{0.27} \\ \hline
PEMS-SF & 267 & \multicolumn{1}{c|}{0.8266} & \multicolumn{1}{c|}{\textbf{3.53}} & 2.13 & \multicolumn{1}{c|}{\textbf{0.8284}} & \multicolumn{1}{c|}{28.08} & \textbf{0.09} \\ \hline
EthanolConcen & 261 & \multicolumn{1}{c|}{\textbf{0.4448}} & \multicolumn{1}{c|}{\textbf{14.59}} & 14.32 & \multicolumn{1}{c|}{0.3612} & \multicolumn{1}{c|}{131.58} & \textbf{0.69} \\ \hline
Heartbeat & 204 & \multicolumn{1}{c|}{0.7414} & \multicolumn{1}{c|}{\textbf{4.57}} & 4.59 & \multicolumn{1}{c|}{\textbf{0.7853}} & \multicolumn{1}{c|}{17.13} & \textbf{0.09} \\ \hline
SelfRegSCP2 & 200 & \multicolumn{1}{c|}{\textbf{0.5833}} & \multicolumn{1}{c|}{\textbf{10.78}} & 9.65 & \multicolumn{1}{c|}{\textbf{0.5833}} & \multicolumn{1}{c|}{50.05} & \textbf{0.22} \\ \hline
NATOPS & 180 & \multicolumn{1}{c|}{0.8944} & \multicolumn{1}{c|}{\textbf{0.6}} & 0.58 & \multicolumn{1}{c|}{\textbf{0.9444}} & \multicolumn{1}{c|}{14.61} & \textbf{0.04} \\ \hline
Libras & 180 & \multicolumn{1}{c|}{0.8667} & \multicolumn{1}{c|}{\textbf{0.36}} & 0.29 & \multicolumn{1}{c|}{\textbf{0.9277}} & \multicolumn{1}{c|}{11.51} & \textbf{0.04} \\ \hline
HandMovement & 160 & \multicolumn{1}{c|}{\textbf{0.4189}} & \multicolumn{1}{c|}{\textbf{3.31}} & 1.7 & \multicolumn{1}{c|}{0.4054} & \multicolumn{1}{c|}{11.29} & \textbf{0.03} \\ \hline
RacketSports & 151 & \multicolumn{1}{c|}{\textbf{0.9078}} & \multicolumn{1}{c|}{\textbf{0.29}} & 0.32 & \multicolumn{1}{c|}{0.8618} & \multicolumn{1}{c|}{11.86} & \textbf{0.03} \\ \hline
Handwriting & 150 & \multicolumn{1}{c|}{\textbf{0.5376}} & \multicolumn{1}{c|}{\textbf{0.81}} & 3.92 & \multicolumn{1}{c|}{0.3752} & \multicolumn{1}{c|}{11.85} & \textbf{0.23} \\ \hline
Epilepsy & 137 & \multicolumn{1}{c|}{0.971} & \multicolumn{1}{c|}{\textbf{0.91}} & 0.93 & \multicolumn{1}{c|}{\textbf{0.9855}} & \multicolumn{1}{c|}{10.52} & \textbf{0.03} \\ \hline
EigenWorms & 128 & \multicolumn{1}{c|}{\textbf{0.8702}} & \multicolumn{1}{c|}{\textbf{107.48}} & 111.42 & \multicolumn{1}{c|}{0.5934} & \multicolumn{1}{c|}{225.71} & \textbf{0.7} \\ \hline
UWaveGesture & 120 & \multicolumn{1}{c|}{\textbf{0.9188}} & \multicolumn{1}{c|}{\textbf{1.21}} & 3.07 & \multicolumn{1}{c|}{0.8906} & \multicolumn{1}{c|}{10.2} & \textbf{0.09} \\ \hline
Cricket & 108 & \multicolumn{1}{c|}{\textbf{1}} & \multicolumn{1}{c|}{\textbf{5.79}} & 4.07 & \multicolumn{1}{c|}{\textbf{1}} & \multicolumn{1}{c|}{32.1} & \textbf{0.1} \\ \hline
DuckDuckGeese & 50 & \multicolumn{1}{c|}{0.5} & \multicolumn{1}{c|}{\textbf{1.59}} & 1.76 & \multicolumn{1}{c|}{\textbf{0.62}} & \multicolumn{1}{c|}{9.46} & \textbf{0.05} \\ \hline
BasicMotions & 40 & \multicolumn{1}{c|}{\textbf{1}} & \multicolumn{1}{c|}{\textbf{0.25}} & 0.27 & \multicolumn{1}{c|}{\textbf{1}} & \multicolumn{1}{c|}{4.45} & \textbf{0.01} \\ \hline
ERing & 30 & \multicolumn{1}{c|}{\textbf{0.9851}} & \multicolumn{1}{c|}{\textbf{0.17}} & 0.7 & \multicolumn{1}{c|}{0.9629} & \multicolumn{1}{c|}{3.17} & \textbf{0.06} \\ \hline
AtrialFibrillation & 15 & \multicolumn{1}{c|}{0.2} & \multicolumn{1}{c|}{\textbf{0.39}} & 0.41 & \multicolumn{1}{c|}{\textbf{0.4}} & \multicolumn{1}{c|}{1.99} & \textbf{0.01} \\ \hline
StandWalkJump & 12 & \multicolumn{1}{c|}{\textbf{0.5333}} & \multicolumn{1}{c|}{\textbf{1.52}} & 1.65 & \multicolumn{1}{c|}{0.3333} & \multicolumn{1}{c|}{14.56} & \textbf{0.09} \\ \hline
\end{tabular}
\end{table}

\subsection{ConvTran vs non-deep learning SOTA Models}
Table~\ref{tab:Conv_HC} compares the performance of ConvTran against three non-deep learning models - ROCKET, HC2, and CIF - on different datasets with varying training sample sizes. The table presents the accuracy of each model on each dataset, with boldface indicating superior accuracy. ``-'' denotes non-runnable methods, either due to computation complexity or inability to handle various length series.

Overall, ConvTran outperforms the non-deep learning models on 19 out of 32 datasets (for the HC2 and CIF models, we only have results for 26 datasets, and ConvTran outperforms the other models in 13 out of the 26). It performs better on datasets with larger training sample sizes, such as InsectWingBeat, while other models perform better on datasets with fewer training samples, such as StandWalkJump, which only has 12 training samples.
Additionally, the table shows that some of the non-deep learning models failed to handle specific datasets due to either computational complexity or the inability to handle varying input series lengths. 
For example, we were not able to run HC2 and CIF on the larger HAR, Ford, and InsectWingbeat datasets due to computational complexity.
They were also not designed to handle varying length time series such as the CharacterTrajectories, SpokenArabicDigits, and JapaneseVowels datasets.

\begin{table}
\centering
\setlength{\tabcolsep}{3.5 pt}
\setlength\extrarowheight{1.5 pt}
\caption{Comparison of ConvTran and Non-Deep Learning Models (ROCKET, HC2, CIF) on Varying Training Sample Sizes. Bold Face Font Indicates Superior Accuracy, '-' Denotes Non-Runnable Methods due to Computation Complexity or Inability to Handle Various Length Series.}
\label{tab:Conv_HC}
\begin{tabular}{|l|c|c|c|c|c|}
\hline
\textbf{Datasets} & \textbf{Train Size} & \textbf{ConvTran} & \textbf{ROCKET} & \textbf{HC2} & \textbf{CIF} \\ \hline \hline
HAR & 41546 & \textbf{0.9098} & 0.8293 & - & - \\ \hline
Ford & 28839 & \textbf{0.7805} & 0.6051 & - & - \\ \hline
InsectWingbeat & 25000 & \textbf{0.7132} & 0.4182 & - & - \\ \hline
PenDigits & 7494 & \textbf{0.9871} & 0.984 & 0.9791 & 0.9674 \\ \hline
SpokenArabicDigits & 6599 & \textbf{0.9945} & 0.9932 & - & - \\ \hline
FaceDetection & 5890 & \textbf{0.6722} & 0.5624 & 0.6603 & 0.6271 \\ \hline
PhonemeSpectra & 3315 & \textbf{0.3062} & 0.1894 & 0.2905 & 0.2654 \\ \hline
LSST & 2459 & 0.6156 & 0.5251 & \textbf{0.6427} & 0.5726 \\ \hline
CharacterTrajectories & 1422 & \textbf{0.9922} & 0.9916 & - & - \\ \hline
FingerMovements & 316 & \textbf{0.56} & 0.55 & 0.53 & 0.52 \\ \hline
MotorImagery & 278 & \textbf{0.56} & \textbf{0.56} & 0.54 & 0.5 \\ \hline
ArticularyWordRecognition & 275 & 0.9833 & \textbf{0.9933} & \textbf{0.9933} & \textbf{0.9833} \\ \hline
JapaneseVowels & 270 & \textbf{0.9891} & 0.9568 & - & - \\ \hline
SelfRegulationSCP1 & 268 & \textbf{0.918} & 0.8601 & 0.8908 & 0.8601 \\ \hline
PEMS-SF & 267 & 0.8284 & 0.8266 & \textbf{1} & \textbf{1} \\ \hline
EthanolConcentration & 261 & 0.3612 & 0.346 & \textbf{0.7719} & 0.7338 \\ \hline
Heartbeat & 204 & \textbf{0.7853} & 0.678 & 0.7317 & 0.7805 \\ \hline
SelfRegulationSCP2 & 200 & \textbf{0.5833} & \textbf{0.5833} & 0.5 & 0.5 \\ \hline
NATOPS & 180 & \textbf{0.9444} & 0.8944 & 0.8944 & 0.8556 \\ \hline
Libras & 180 & 0.9277 & 0.8667 & \textbf{0.9333} & 0.9111 \\ \hline
HandMovementDirection & 160 & 0.4054 & 0.4189 & 0.473 & \textbf{0.5946} \\ \hline
RacketSports & 151 & 0.8618 & \textbf{0.9078} & \textbf{0.9078} & 0.8816 \\ \hline
Handwriting & 150 & 0.3752 & 0.5376 & \textbf{0.5482} & 0.3565 \\ \hline
Epilepsy & 137 & 0.9855 & 0.971 & \textbf{1} & 0.9855 \\ \hline
EigenWorms & 128 & 0.5934 & 0.8702 & \textbf{0.9466} & 0.916 \\ \hline
UWaveGestureLibrary & 120 & 0.8906 & 0.9188 & \textbf{0.9281} & 0.925 \\ \hline
Cricket & 108 & \textbf{1} & \textbf{1} & \textbf{1} & 0.9861 \\ \hline
DuckDuckGeese & 50 & \textbf{0.62} & 0.5 & 0.56 & 0.44 \\ \hline
BasicMotions & 40 & \textbf{1} & \textbf{1} & \textbf{1} & \textbf{1} \\ \hline
ERing & 30 & 0.9629 & 0.9593 & \textbf{0.9889} & 0.9815 \\ \hline
AtrialFibrillation & 15 & \textbf{0.4} & 0.1333 & 0.2667 & 0.3333 \\ \hline
StandWalkJump & 12 & 0.3333 & \textbf{0.5333} & 0.4667 & 0.4 \\ \hline \hline
\textbf{Wins or Draw} & - & \textbf{19} & \textbf{7} & \textbf{13} & \textbf{4} \\ \hline
\end{tabular}
\end{table}

\end{appendices}
\end{document}